\documentclass[letterpaper, 10 pt, journal, twoside]{IEEEtran}
\ifCLASSINFOpdf
\else
\fi
\usepackage{color}
\usepackage{amsmath}
\usepackage{graphicx}
\usepackage{subfigure}
\usepackage{makecell}
\usepackage{booktabs}
\usepackage{gensymb}
\usepackage{hyperref}
\usepackage{threeparttable}
\usepackage{multicol}
\usepackage{url}

\begin{document}
\title{Efficient learning of goal-oriented push-grasping synergy in clutter}

% \author{Michael~Shell,~\IEEEmembership{Member,~IEEE,}
%         John~Doe,~\IEEEmembership{Fellow,~OSA,}
%         and~Jane~Doe,~\IEEEmembership{Life~Fellow,~IEEE}% <-this % stops a space
% \thanks{M. Shell was with the Department
% of Electrical and Computer Engineering, Georgia Institute of Technology, Atlanta,
% GA, 30332 USA e-mail: (see http://www.michaelshell.org/contact.html).}% <-this % stops a space
% \thanks{J. Doe and J. Doe are with Anonymous University.}% <-this % stops a space
% \thanks{Manuscript received April 19, 2005; revised August 26, 2015.}}
\author{Kechun Xu$^{1}$, Hongxiang Yu$^{1}$, Qianen Lai$^{1}$, Yue Wang$^{1}$, Rong Xiong$^{1}$
%First A. Author$^{1}$, Second B. Author$^{2}$, and Third C. Author$^{1}$%
\thanks{Manuscript received: February, 24, 2021; Revised May, 28, 2021; Accepted June, 15, 2021.}%Use only for final RAL version
\thanks{This paper was recommended for publication by Editor Markus Vincze upon evaluation of the Associate Editor and Reviewers' comments.)} %Use only for final RAL version

\thanks{$^{1}$Kechun Xu, Hongxiang Yu, Qianen Lai, Yue Wang, Rong Xiong are with the State Key Laboratory of Industrial Control Technology and Institute of Cyber-Systems and Control, Zhejiang University, Hangzhou, China. Yue Wang is the corresponding author,
        {\tt\footnotesize wangyue@iipc.zju.edu.cn}}%
        
% \thanks{$^{1}$First Author and Third Author are with School of Engineering, Robotics Department, University of Somewhere, Someland
%         {\tt\footnotesize first.author@papercept.net}}%

% \thanks{$^{2} $SecondAuthor is with School of Engineering, Automation Department, University of Anywhere, Anyland
%         {\tt\footnotesizel second.author@papercept.net}}%
\thanks{Digital Object Identifier (DOI): see top of this page.}
}

\markboth{IEEE Robotics and Automation Letters. Preprint Version. Accepted June, 2021}
{Xu \MakeLowercase{\textit{et al.}}: Efficient learning of goal-oriented push-grasping synergy in clutter} 




\maketitle

\begin{abstract}
We focus on the task of goal-oriented grasping, in which a robot is supposed to grasp a pre-assigned goal object in clutter and needs some pre-grasp actions such as pushes to enable stable grasps. {However, in this task, the robot gets positive rewards from environment only when successfully grasping the goal object. Besides, joint pushing and grasping elongates the action sequence, compounding the problem of reward delay. Thus, sample inefficiency remains a main challenge in this task.} In this paper, a goal-conditioned hierarchical reinforcement learning formulation with high sample efficiency is proposed to learn a push-grasping policy for grasping a specific object in clutter. In our work, sample efficiency is improved by two means. First, we use a goal-conditioned mechanism by goal relabeling to enrich the replay buffer. Second, the pushing and grasping policies are respectively regarded as a generator and a discriminator and the pushing policy is trained with supervision of the grasping discriminator, thus densifying pushing rewards. To deal with the problem of distribution mismatch caused by different training settings of two policies, an alternating training stage is added to learn pushing and grasping in turn. A series of experiments carried out in simulation and real world indicate that our method can quickly learn effective pushing and grasping policies and outperforms existing methods in task completion rate and goal grasp success rate by less times of motion. Furthermore, we validate that our system can also adapt to goal-agnostic conditions with better performance. Note that our system can be transferred to the real world without any fine-tuning. Our code is available at {\href{https://github.com/xukechun/Efficient_goal-oriented_push-grasping_synergy}{https://github.com/xukechun/Efficient$\_$goal-oriented$\_$push-grasping$\_$synergy}}
\end{abstract}

% \begin{IEEEkeywords}
% IEEE, IEEEtran, journal, \LaTeX, paper, template.
% \end{IEEEkeywords}

% \fi
\IEEEpeerreviewmaketitle

\section{Introduction}
% for IEEE journal papers produced under \LaTeX\ using
% IEEEtran.cls version 1.8b and later.
\IEEEPARstart{G}{rasping} plays an essential role in robot manipulation since it is a basic action for lots of complex tasks e.g. table organization \cite{batra2020rearrangement}. 
Among these tasks, the robot usually faces a cluttered scene, where the tight packing around the goal objects may seriously degenerate the successful grasping. Inspired by the human behavior, the synergy between pushing and grasping becomes a solution to deal with such scenarios. By pushing the clutter, there can be space around the goal object for robot gripper insertion. To bring the skill to the robot, the vital problem is to develop the manipulation policy for synergy so that grasping with high success rate can be achieved using less steps of pushing.

There are several works \cite{zeng2018learning}\,\cite{boularias2015learning}\,\cite{deng2019deep}\,\cite{chen2020combining}\,\cite{HuaHanBouYu21ICRA} establishing the learning strategies to develop the synergies between pushing and grasping. \cite{zeng2018learning} learns pushing and grasping policies equally using two parallel networks for push and grasp. \cite{deng2019deep} and \cite{chen2020combining} design a hierarchical framework by evaluating whether {the robot} is ready to grasp after current push step. One disadvantage of these methods is that the grasping object cannot be designated, thus only applicable to goal-agnostic tasks. Simply adapting these methods to goal-oriented tasks is to apply a mask of goal object on the grasping map, which, however, showing unsatisfactory results \cite{yang2020deep}.

{In contrast, goal-oriented grasping task, which is to grasp a pre-assigned goal object from a cluttered scene, is a more challenging problem, which needs to effectively incorporating the goal object information in push-grasping synergy}. \cite{dogar2013physics} uses physics-based analysis to compute pushing and grasping with a known 3D model of the goal object. \cite{marios2019robust} and \cite{sarantopoulos2020split} employ sparse rewards for training. Hence the robot gets positive rewards only when the goal object is singulated. Note that it is more difficult for the robot to acquire a positive reward from the environment in goal-oriented tasks compared with goal-agnostic tasks, causing low sample efficiency for training. 
{To deal with this problem, \cite{yang2020deep} develops denser rewards by assigning rewards for pushing totally with handcraft to encourage moving objects and removing clutter.}
But \cite{danielczuk2018linear} argues that the relationship between pushing and grasping is complex and coupled, leading to the handcrafted design of an optimal pushing reward function very difficult. 

\begin{figure}[t]
  \centering
  \includegraphics[width=0.8\linewidth]{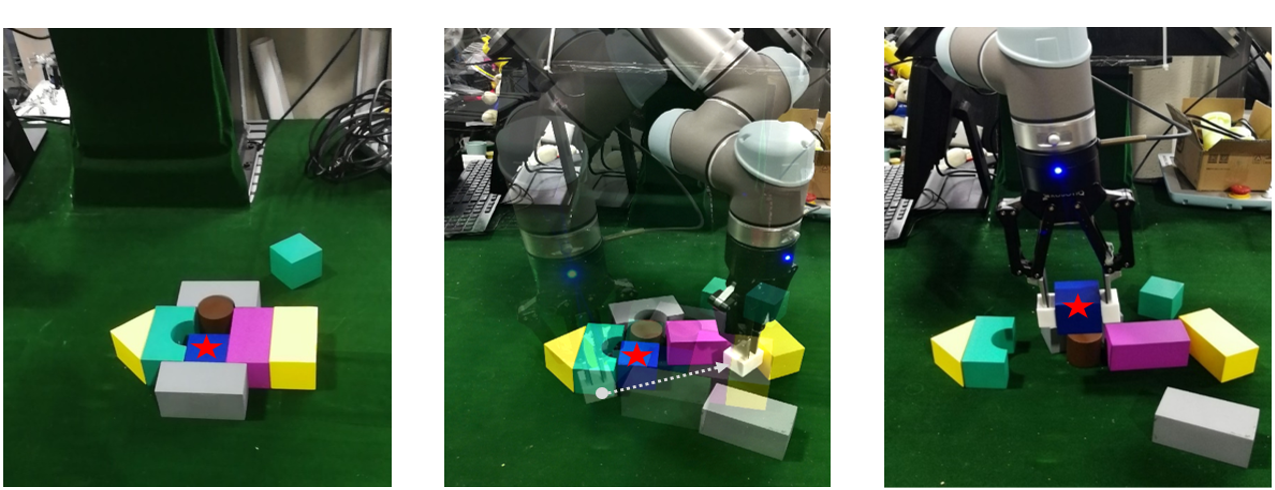}
  \vspace{-0.2cm}
  \caption{An example scenario ({left}) of goal-oriented grasping tasks, in which the robot is supposed to grasp the object labeled with a star in clutter. The tight packing around the goal object degenerates the successful grasping thus synergy between grasping and pre-actions like pushing is needed. Our system is capable of planning push actions (middle) to isolate the goal object and finally executing a stable grasp action (right) to pick it.}
  \label{fig:example_action}
  \vspace{-0.7cm}
\end{figure}

In this paper, we propose an efficient learning-based push-grasping synergy policy for goal-oriented grasping tasks (Fig. \ref{fig:example_action}) by addressing the problems of sample inefficiency and sparse rewarding. We state the task as a goal-conditioned hierarchical reinforcement learning problem, and learn the model from scratch. By leveraging a goal-conditioned learning mechanism, denser rewards can be acquired to accelerate the learning. Specifically, if successfully grasping an object other than the designated one, we relabel the goal of corresponding experience in the replay buffer to the grasped one to improve the sample efficiency. Then, we design rewards for pushing by applying adversarial training inversely to build the synergy between pushing and grasping. With grasping as a discriminator, the adversarial training encourages the robot to push the clutter and generate a scene where the goal object is graspable with high probability. Thus the efficiency can be improved. As a whole, we train the grasp and push nets in an alternating way to relieve the problem of distribution mismatch, finally developing the coordination between the two actions. Experiments conducted in simulation and real-world environments demonstrate that our system can achieve better performances in the goal-oriented grasping task with higher task completion and grasp success rate in less steps (Sec. \ref{experiment}). Note that our system is transferred from simulation to the real world {without} any additional data collection for fine-tuning. {Moreover, our study on goal-oriented grasping is also applicable for goal-agnostic tasks by removing the target information (validated in Sec. \ref{experiment})}. To summarize, the main contributions of this paper are:
\begin{itemize}
\item We propose a relabeling driven goal-conditioned strategy and an adversarial training based dense reward formulation to improve the sample efficiency for learning goal-oriented synergy between pushing and grasping.
\item We propose an additional alternating training stage by learning the grasp network and push network in turn to improve the system performance against the distribution mismatch problem.
\item We evaluate the learned system not only on both simulated random and designed cluttered scenes, but also on real world scenarios without extra fine-tuning, of which the results validate the effectiveness.
\end{itemize}


%%%%%%%%%%%%%%%%%%%%%%%%%%%%%%%%%%%%%%%%%%%%%%%%%%%%%%%%%%%%%%%%%%%%%%%%%%%%%%%%
\section{RELATED WORK}
\subsection{Grasping}
Robotic grasping techniques have been well studied for decades. Such methods can be divided into two categories: analytical and data-driven \cite{bohg2013data}. Classic analytical methods require 3D models of known objects to find stable force closures for grasping \cite{liang2019pointnetgpd}\,\cite{rodriguez2012caging}\,\cite{sahbani2012overview}. However, most of {the} time it is difficult to get accurate models of novel objects. Recent data-driven approaches \cite{choi2018learning}\,\cite{mahler2017dex} focus on directly mapping from visual observations to actions. But most of them assume the scene is scattered enough in which objects are well isolated. Learning for robotic grasping in cluttered environments is studied in \cite{kalashnikov2018scalable}\,\cite{ten2018using}\,\cite{pinto2016supersizing}\,\cite{mahler2017learning}. Unfortunately, picking objects in a dense clutter with only grasp actions is insufficient \cite{ten2018using}. Pre-grasp manipulations such as pushing might be necessary for singulation.
\subsection{Pushing assisted grasping}
Synergies between pushing and grasping have been explored in \cite{zeng2018learning}\,\cite{boularias2015learning}\,\cite{deng2019deep}\,\cite{chen2020combining}\,\cite{HuaHanBouYu21ICRA}, which aim to rearrange cluttered objects by push actions to enable future grasps. Zeng {\it et al.} \cite{zeng2018learning} proposes a model-free deep reinforcement learning framework to learn joint policies of pushing and grasping with parallel architecture. \cite{deng2019deep} and \cite{chen2020combining} evaluate whether {the goal object} is ready for grasping by analytical metrics or neural networks, otherwise executing push actions to rearrange the clutter. Huang {\it et al.} \cite{HuaHanBouYu21ICRA} predicts push actions as well as states after pushing so that it is more intuitive to judge whether it is time to grasp.

Compared with above goal-agnostic methods, goal-oriented push-grasping in cluttered scene is much less explored, except for \cite{yang2020deep}\,\cite{dogar2013physics}\,\cite{marios2019robust}\,\cite{sarantopoulos2020split}. \cite{dogar2013physics} requires the 3D model of the goal object and uses a physics-based analysis of pushing to compute the motion of each object in the scene. \cite{marios2019robust} learns a pushing policy via Q-learning to singulate the goal object from its surrounding clutter. \cite{sarantopoulos2020split} proposes a split DQN framework which speeds up training process compared with \cite{marios2019robust}. \cite{yang2020deep} enables the goal object to be invisible in the initial state and uses a Bayesian-based policy to search it. 
{Besides, other works on learning pushing networks for singulation in the context of mechanical search can also be regarded as push-assisted grasping methods in a broad sense. \cite{eitel2020learning} uses a push proposal network to rank pushes sampled upon object segments. The label of a push action is determined by a user who assesses the outcome of the action. \cite{novkovic2020object} proposes a RL approach for active and interactive perception to search a goal object in clutter, in which rewards for actions are handcrafted based on exploration and detection results. \cite{Kurenkov2020VisuomotorMS} applies a teacher-guided strategy to train the pushing policy with pushing rewards based on the change of occlusion.} 

Analogous to above goal-oriented methods \cite{yang2020deep}\,\cite{marios2019robust}\,\cite{sarantopoulos2020split} \cite{Kurenkov2020VisuomotorMS}, our approach is data-driven. There are two main differences. {First, methods in \cite{yang2020deep}\,\cite{marios2019robust}\,\cite{sarantopoulos2020split} design the pushing reward function totally by handcraft, which might require many iterations of tuning \cite{ratner2018simplifying} and lack of generalization to new scenarios \cite{hadfield2017inverse}. In contrast, a data-driven reward will automatically optimize itself through the robot’s experience to learn insight beyond human intuition, which is also more time-saving and effective. However, although totally handcrafted rewards are limited and defective, they can serve as effective approximation of the true rewards \cite{hadfield2017inverse} to effectively guide the agent. Our approach aims to get pushing rewards mainly from a data-driven neural network which predicts the probability of successful grasps, to learn from posterior experience, combined with proxy handcrafted pushing rewards as important prior knowledge. Second, previous goal-oriented grasping or singulation works do not leverage experience of non-goal objects' grasping or singulation, thus requiring more samples to train the policy. {\cite{yang2020deep} needs around 2k training episodes to build its push-grasping coordinator with 80$\%$ grasp success rate. \cite{Kurenkov2020VisuomotorMS} converges
to high improvements in graspability within 8k training steps. Other goal-agnostic works, like VPG\,\cite{zeng2018learning}, spend more than 2.5k training episodes to reach 80$\%$ grasp success rate}. Inspired by Hindsight Experience Replay (HER) \cite{andrychowicz2017hindsight}, a goal relabeling algorithm, we use a goal-conditioned learning mechanism to relabel the pushing and grasping experience with grasping goals to improve the sample efficiency, in which we extend the goal of pushing as its corresponding grasping goals.}

%%%%%%%%%%%%%%%%%%%%%%%%%%%%%%%%%%%%%%%%%%%%%%%%%%%%%%%%%%%%%%%%%%%%%%%%%%%%%%%%
\section{METHOD}
\label{method}

\begin{figure*}[t]
  \centering
  \includegraphics[width=0.85\textwidth]{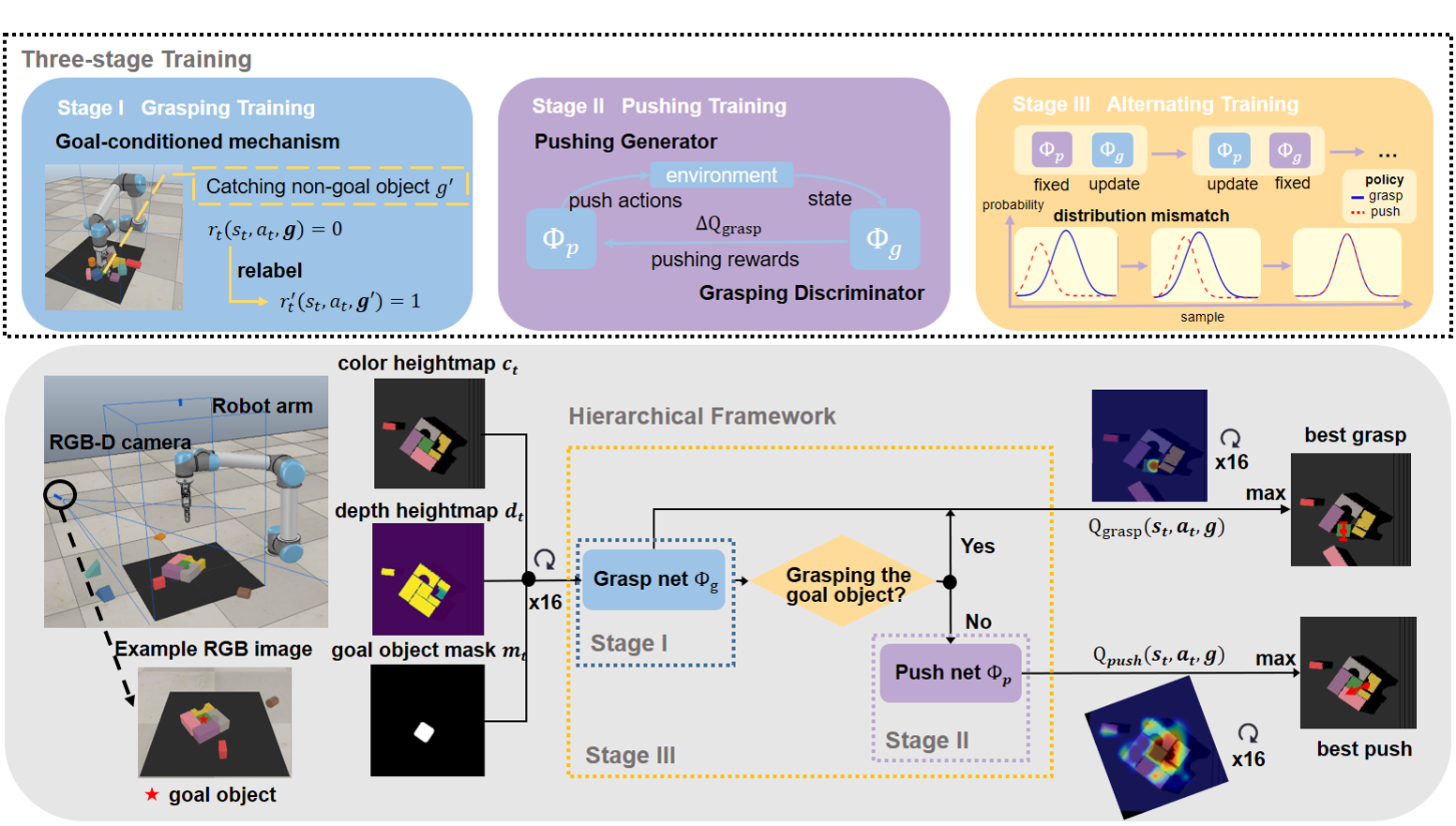}
  \vspace{-0.2cm}
  \caption{{\bf Overview} of our hierarchical framework and three-stage training approach. We fix a camera to capture RGB-D images of the workspace and construct visual heightmaps and the goal mask by orthographically projecting images in the gravity direction. The grasp net serves as a discriminator that evaluates whether is ready to grasp the goal object to choose between pushing and grasping. Color heightmaps, depth heightmaps and masks of the goal object are fed into the grasp and push nets, which predict pixel-wise grasp and push Q maps. Example of Q maps and predicted actions are also presented in this figure. Push is parameterized by a start position and a pushing orientation with a fixed pushing length, while grasp is defined by a middle position and an orientation of the parallel jaw of the gripper with a fixed width.}
  \label{fig:overview}
  \vspace{-0.6cm}
\end{figure*}

{
% We model the goal-oriented push-grasping problem as a goal-conditioned MDP: given a state $s_t$ and a goal object representation $g$ at time $t$, the robot executes an action $a_t$ according to a goal-conditioned policy $\pi(s_t|g)$, then turns to a new state $s_{t+1}$ while getting an immediate reward $R(s_t,a_t,g)$. The expected sum of future rewards is given by $R_{t}=\sum_{i=t}^{\infty} \gamma R\left(s_{i}, a_i, g\right)$, where $\gamma$ is the discount factor. In our work, DQN \cite{mnih2015human} is employed to learn the policy $\pi(s_t|g)$ that chooses the $a_t$ with the highest action-value of the Q function ({\it i.e.} $Q_\pi (s_t,a_t,g)$), which measures the expected reward of taking action $a_t$ in state $s_t$ for goal $g$ at time $t$. Our target is to minimize the error between $Q_\pi (s_t,a_t,g)$ and the target value $y_t$ which is given by the following formula.
% \begin{equation}
% \begin{aligned}
% y_{t}&=R\left(s_{t}, a_{t}, g\right)+
% \gamma Q_{\pi}\left(s_{t+1}, a_{t+1}, g\right)
% \end{aligned}
% \end{equation}
% where $a_{t+1}=\operatorname{argmax}_{a^{\prime}}Q_{\pi}\left(s_{t+1}, a^{\prime}, g\right)$, in which $a^{\prime}$ is the set of actions ({\it i.e.} push and grasp).
We model the goal-oriented push-grasping problem as a goal-conditioned MDP, which is very similar to that in VPG \cite{zeng2018learning}, with a new symbol for the goal object representation as $g$. In this formulation, the symbol definitions of policy, reward and Q function are all conditioned on the goal $g$, which is defined as $\pi(s_t|g)$, $R(s_t,a_t,g)$ and $Q_\pi (s_t,a_t,g)$ respectively.
}

As shown in Fig. \ref{fig:overview}, we fix a camera to capture RGB-D images of the workspace. Then we represent each state $s_t$ with a color heightmap $c_t$ as well as a depth heightmap $d_t$, obtained by projection of the original RGB-D images in the gravity direction \cite{zeng2018learning}. We represent the goal object as a mask heightmap $m_t$, which indicates the positions occupied by the object in the workspace. {These heightmap representations are rotated by 16 angles with a stationary step size of {22.5\degree} before fed into networks to account for different grasp rotations w.r.t the z-axis}.

We formulate the task of pushing and grasping in cluttered environments as a hierarchical reinforcement learning problem. For high-level control, our grasp net $\phi_g$ serves as a discriminator that scores current state for goal object grasping. {If the score for grasping the goal object exceeds a threshold, the robot will execute the grasp action with maximal Q value}. Otherwise low-level control will be activated: the robot executes push actions to transfer the state until it becomes graspable for the discriminator. Both push and grasp nets ({\it i.e.} $\phi_p$, $\phi_g$) take as input the heightmap representations of state $s_t$ as well as the masks of goal object, and output dense pixel-wise maps of Q values with the same image size and resolution as that of heightmap representations. Every Q value at a pixel represents the future expected reward of executing primitive at the corresponding 3D location whose depth component is computed from the depth heightmap. As for our action definition, push is parameterized by a start position and a pushing orientation with a fixed pushing length, while grasp is defined by a middle position and an orientation of the parallel jaw of the gripper with a fixed width.

To train this hierarchical framework, a three-stage training approach is proposed as follows.

{\bf Stage \uppercase\expandafter{\romannumeral1}: Goal-conditioned Grasping Training.} In the first stage, we train a goal-conditioned grasp net. To improve the sample efficiency, a goal-conditioned learning mechanism is introduced to relabel the goal of grasping experience to the grasped one. {After enough episodes, expected Q values of successful grasps are stable above a specific value $Q_g^*$ (whose determination method will be discussed in Sec \ref{simulated_experiment} in detail), which is set as a threshold that distinguishes appropriate states for goal object grasping}. 

{\bf Stage \uppercase\expandafter{\romannumeral2}: Goal-conditioned Pushing Training.} In this stage, we focus on training a push net, and parameters of the pre-trained grasp net are fixed. Push is considered effective only when assisting future grasps, thus our pushing reward function is designed based on the grasp net by utilizing adversarial training inversely. Positive rewards will be obtained if pushes improve the graspable probability predicted by the grasp net. Also we use the goal-conditioned mechanism which relabels the goal of grasping and sequential pushing experience with the grasped one to improve the sample efficiency.

{\bf Stage \uppercase\expandafter{\romannumeral3}: Alternating Training.} In the previous stage, we just consider the grasp net as an expert and fix its network parameters when training the push net. However, the former grasping net is trained in a relatively scattered environment where there is little occlusion around the goal object with only grasp actions. In this stage, we aim to further improve the performance of our pushing and grasping policies in cluttered environments by alternately training to relieve the the problem of distribution mismatch.

\subsection{Goal-conditioned Grasping Training}
\label{grasp training}
In this stage the robot only executes grasp actions, and each grasp corresponds to an episode. To reduce the influence of occlusion, we create a relatively scattered scene where the goal object is well isolated by randomly dropping $a$ objects in the workspace. We define the grasping reward function as:
\begin{equation}
\label{eqn:grasp reward}
R_{g}=\left\{\begin{array}{ll}
1, & \text { if\;grasping\;goal\;object\;successfully } \\
0, & \text { otherwise }
\end{array}\right.
\end{equation}

There are essential differences between the two cases where the robot gets no reward: failed grasps are totally harmful to the grasping performance, while the wrong object catches can serve as treasured experiences to improve the sample efficiency. {Under this motivation, we use a relabeling driven goal-conditioned strategy.} When the goal object is occluded by others, the robot is likely to accidentally grasp a non-goal object $g^{\prime}$. If that happens, we relabel the goal $g$ with $g^{\prime}$ in the transition tuple $(s_{t},a_{t},r_{t},s_{t+1},g)$ as $(s_t, a_t, r_t^{\prime}, s_{t+1}, g^{\prime})$, where $r_t^{\prime}:=R(s_t,a_t,g^{\prime})$ and save this experience into the replay buffer to train our policy. By leveraging these wrong-object catching samples, we improve the sample efficiency thus speeding up the training.
% Our goal-conditioned training mechanism is shown in Algorithm \ref{alg:1}.

% \begin{algorithm}[htb]
% \caption{Goal-conditioned Grasping Training}
% \label{alg:1}
% \begin{algorithmic}
% \STATE {\bf Given:} a goal set $G$;
% \STATE Initialize a goal-conditioned policy $\pi_g(s_t|g)$, where $s_t$ is state at time $t$ and $g$ is a representation of goal;
% \STATE Initialize experience replay buffer $B_g$;
% \FOR{each episode $i$}
%     \STATE Sample a goal $g_i \in G$;\
%     \STATE Choose an action $a_{t}$ from policy: $a_{t} \leftarrow \pi_g(s_t|g_i)$; \
%     \STATE Execute action $a_{t}$, tranfer to state $s_{t+1}$, and get reward $r_t:=R_g(s_t,a_{t},g_i)$;\
%     \STATE Save the transition $(s_t, a_t, r_t, s_{t+1},g_i)$ into $B_g$;\
%     \IF{catching non-goal object $g_i^{\prime}$}
%         \STATE Save the transition $(s_t, a_t, r_t^{\prime}, s_{t+1}, g_i^{\prime})$ into $B_g$, where $r_t^{\prime}:=R_g(s_t,a_{t},g_i^{\prime})$;
%     \ENDIF
%     \STATE Sample data from $B_g$ to train $\pi_g(s_t|g)$;
% \ENDFOR

% \end{algorithmic}
% \end{algorithm}

After enough training episodes, Q values of successful grasps are stable above a specific value $Q_g^*$. In the following stages, we consider the current state to be ready for goal object grasping when the highest predicted Q value within the goal mask exceeds $Q_g^*$.

\subsection{Goal-conditioned Pushing Training}
For this stage our target is to learn pushes to enable future grasps using as few steps as possible. We limit the push number up to five in an episode with a grasp in the end. Besides, if the highest Q value within the goal mask exceeds the threshold $Q_g^*$, the agent will immediately execute the terminal grasp action. During one episode $b$ different objects are randomly dropped in the scene to create clutter.

Pushes are favorable when enabling later grasps. However, the interaction between pushing and grasping is complex and coupled, thus difficult to model pushing rewards totally by handcraft \cite{danielczuk2018linear}. To face this challenge, we train our pushing policy under the guidance of the grasp net by applying adversarial training inversely. Since grasp is a one-step action, the grasp net can evaluate graspable probability of the current state with its predicted grasp q values. We regard the grasp net as a discriminator that distinguishes a good state for goal object grasping with a Q value threshold $Q_g^*$ which infers a high graspable probability. Correspondingly, push net is modeled as a generator that aims to fool the grasping discriminator by predicting push actions which rearrange the clutter to a state with maximal grasp q value of the goal object exceeding $Q_g^*$. We define the state transition function as $\mathcal{T}$, with $s_{t+1}=\mathcal{T}(s_t,a_t)$. Then our training objective is formulated as:
\begin{equation}
\min _{{\phi}_{{p}}} V\left(D_{g}, G_{p}\right)=E\left[\log \left(1-D_{g}\left(G_{p}\left(s_{t}, g\right)\right)\right)\right]
\end{equation}
\begin{equation}
s_{t+1}=G_{p}\left(s_{t}, g\right)=\mathcal{T}\left(s_{t},\operatorname{argmax}_{a_{{t}}}\phi_{p}\left(s_{t}, a_{{t}}, g\right)\right)
\end{equation}
\begin{equation}
D_{g}(s_{t})=\left\{\begin{array}{ll}
1, & \delta>0 \\
-\delta, & \delta<0
\end{array}\right.
\end{equation}
where $\delta=\max _{a_{{t}}} \phi_{g}\left(s_{t}, a_{{t}}, g\right)-Q_{g}^{*}$. In other words, the target of each push action is to continually increase the maximal grasp Q value of the goal object until exceeding $Q_{g}^{*}$. Based on above analysis, we model our pushing reward function as: 
\begin{equation}
R_{p}=\left\{\begin{array}{ll}
0.5, & {Q}_{{g}}^{\text {improved }}>0.1 \text { and change detected } \\
-0.5, & \text { no change detected } \\
0, & \text { otherwise }
\end{array}\right.
\end{equation}
where $Q_g^{\text {improved }}=Q_g^{\text {after\;pushing }}-Q_g^{\text {before\;pushing }}$. And change detected means the surrounding of the goal object is changed and $o_g^{\text{ decreased}}>0.1$, where $o_g^{\text{ decreased}}$ represents the decrease of occupy ratio surrounding the goal object and is computed from the depth heightmap. As for grasp actions, we follow the reward function (\ref{eqn:grasp reward}) and our reward scheme is the combination of $R_g$ and $R_p$.

Same as grasping training, a goal-conditioned mechanism is introduced for pushing by goal relabeling. If grasping a non-goal object when the goal object is occluded by others, goals of the sequential push actions in the whole episode will be set to the grasped object which means such push sequence is beneficial for its grasping.

\subsection{Alternating Training}
In previous stages our grasping discriminator is trained in a scattered scene only with grasp actions while pushing generator is learned in a relatively cluttered scene with push-grasping collaboration, which may lead to the distribution mismatch problem. That is, the grasping discriminator may fail to distinguish some novel scenes for grasping in push-grasping collaboration. Moreover, the training of grasp net has a great impact on the push net since our pushing reward function is based on the grasp net. To deal with this problem, we further train pushing and grasping policies based on networks $\phi_p$ and $\phi_g$ trained in previous stages. In this stage, grasping and pushing policies are alternately trained with the other net’s parameters fixed. By alternating training, pushing and grasping policies restrict the positive samples of each other. During this training process, we also match the data distribution of the two policies for their synergy application. Therefore, the coordination performance of pushing and grasping is further improved.

Episode setting is the same for pushing and grasping training with $c$ objects randomly dropped in the workspace. For each episode, the robot executes push actions until the maximal Q value within the goal mask exceeds $Q_g^*$. The reward functions remain the same as the former stages.

\subsection{Implementation Details}

In the first two stages, $a=5$ objects are randomly dropped into the workspace for grasping training and $b=10$ objects for pushing training. For the last stage, since our pushing policy has been former trained in clutter with push-grasping collaboration, we train only 500 episodes for pushing while 1000 for grasping, and both with $c=10$ objects in the workspace. Besides, we set the grasp Q value threshold $Q_g^*=1.8$. {Also, since perception is not the main concern of our method, we simplify the perception problem of goal object masks. In our work, goal object masks are obtained by projection of existing 3D models in simulation and color segmentation in the real world. If the goal object is invisible, then all pixels in the mask heightmap are of the value 0.}

Both push and grasp nets ({\it i.e.} $\phi_p$ and $\phi_g$) share the same network architecture, which involves three parallel 121-layer DenseNets \cite{huang2017densely} pretrained on ImageNet \cite{deng2009imagenet} to extract concatenated features of color heightmaps, depth heightmaps and goal masks, followed by a FCN \cite{long2015fully} which contains two $1 \times 1$ convolutional layers with ReLU \cite{nair2010rectified} and batch normalization \cite{ioffe2015batch}, and then bilinearly upsampled. {Both of them are trained using the same loss function as VPG \cite{zeng2018learning}.} At each iteration, gradients are only passed through the single pixel where the action was executed and other pixels backpropagate with 0 loss \cite{zeng2018learning}.

% Both of them are trained using Huber loss at every iteration $i$ :
% \begin{equation}
% \mathcal{L}_{i}=\left\{\begin{array}{cc}
% \frac{1}{2} e_i^2, & \text { if } e_i<1 \\
% e_i-\frac{1}{2}, & \text { otherwise }
% \end{array}\right.
% \end{equation}
% where $e_i=\left|Q_\pi^{\theta_{i}}\left(s_{i}, a_{i}, g\right)-y_i^{\theta_{i}^{-}}\right|$, $\theta_i$ are parameters of $\phi_p$ and $\phi_g$ at iteration $i$ and $\theta_i^-$, the target network parameters, are held fixed between iterations. At each iteration, gradients are only passed through the single pixel where the action was executed and other pixels backpropagate with 0 loss \cite{zeng2018learning}.

We train the networks with Adam optimizer, using fixed learning rates $10^{-4}$, weight decay $2^{-5}$, and betas $(0.9, 0.99)$. We use $\epsilon$-greedy as our exploration strategy, and $\epsilon$ is initialized as 0.5 then annealed to 0.1. Our future discount $\gamma$ is set as a constant at 0.5.

%%%%%%%%%%%%%%%%%%%%%%%%%%%%%%%%%%%%%%%%%%%%%%%%%%%%%%%%%%%%%%%%%%%%%%%%%%%%%%%%
\section{EXPERIMENT}
\label{experiment}

In this section, we carried out a series of experiments to evaluate our system. The goals of the experiments are: 1) to demonstrate our training approach is capable of speeding up the training process effectively. 2) to indicate that our policy is effective and efficient for the task of goal-oriented push-grasping with unseen random and challenging arrangements. 3) to investigate whether our approach can extend to goal-agnostic conditions. 4) to test that our system can successfully transfer from simulation to the real world without any fine-tuning. 

We compare the performance of our system to the following baseline approaches:

{\bf Grasping-only} is a greedy deterministic grasping policy using a single FCN network supervised with binary classification (from trial and error) for grasping. Under this policy agent executes grasping only. In our experiment, the network is replaced by ours. For fair comparisons, this policy is trained with the same simulation environments.

{\bf Goal-conditioned VPG} is an extended version of VPG \cite{zeng2018learning} by adding the goal mask as input to learn goal-oriented grasping and pushing policies. {VPG is a method which predicts Q maps of pushing and grasping for goal-agnostic tasks with parallel architecture. Pushes and grasps with maximal Q value within the goal mask will be executed.}

{\bf Grasping the Invisible} is a target-oriented approach which chooses to push or grasp according to a binary classifier, and applies DQN to train a push-grasping policy \cite{yang2020deep}. A segmentation mechanism is used to detect whether the target is visible so as to switch between exploration and coordination. That is, if the target is {invisible}, the agent executes pushes to explore it and when the target can be seen pushes will coordinate with grasps to catch it.

\subsection{Evaluation Metrics}
We evaluate the methods with a series of test cases where the robot must break dense clutter to pick the goal object. {Each test contains $n$ runs and performances are measured with 3 metrics similar to those in VPG \cite{zeng2018learning} but with goal-oriented definitions:}

\begin{itemize}

\item {\bf Completion}: the average percentage of completion rate over $n$ test runs. {If the policy can pick up the goal object without consecutive 10 fail attempts in grasping the goal object in a run, then the task is considered completed.}
\item {\bf Grasp success rate}: the average percentage of goal grasp success rate per completion.
\item {\bf Motion number}: the average motion number per completion. Motion number reflects action efficiency especially pushing efficiency.
\end{itemize}
\vspace{-0.1cm}
\begin{figure}[t]
  \centering
  \includegraphics[width=0.88\linewidth]{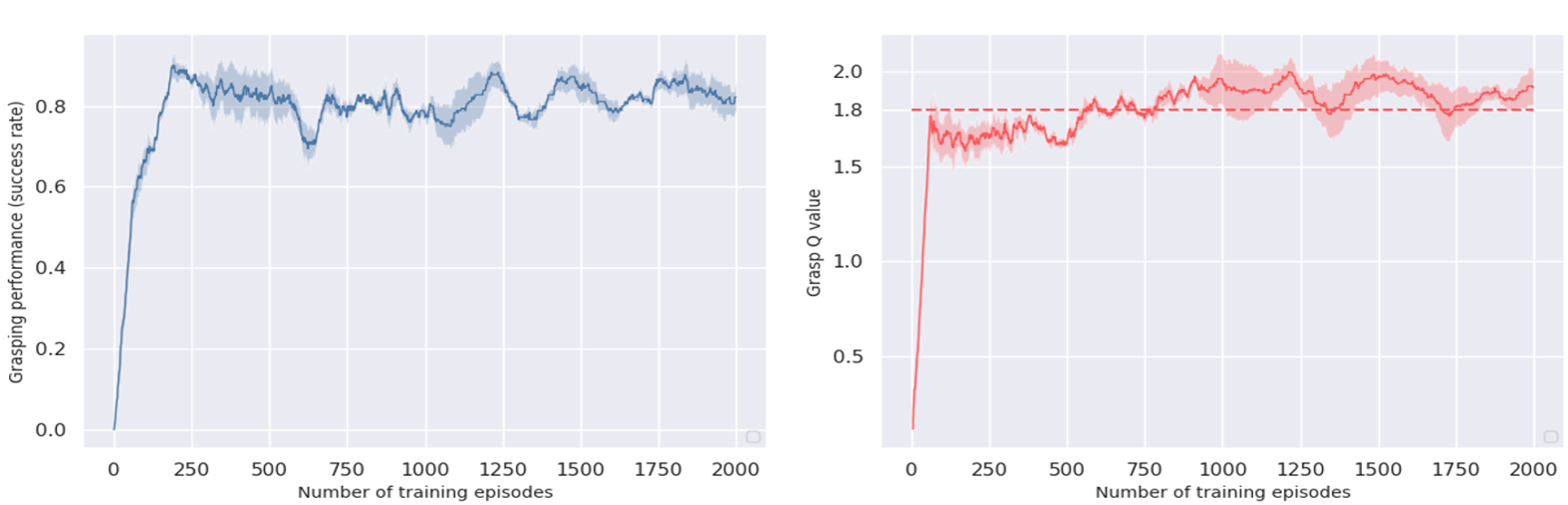}
  \vspace{-0.2cm}
  \caption{Training performance of grasp net. After around 600 training episodes, grasp success rate (left) of goal object becomes stable above 80$\%$ while grasp Q-value (right) stays stably above $Q_g^*=1.8$.}
  \label{fig:grasp_performace}
  \vspace{-0.7cm}
\end{figure}

\subsection{Simulation Experiments}
\label{simulated_experiment}
Our simulation environment is kept the same with \cite{zeng2018learning} for fair comparison, which mainly involves a UR5 arm with an RD2 gripper in V-REP \cite{rohmer2013v}. 

{\bf Parameter Verification.} Our first experiment verifies the threshold $Q_g^*$ in Sec \ref{method}. We report the training performance of grasp net in Fig. \ref{fig:grasp_performace}, where the grasping performance is measured by the percentage of goal grasp success rate over the last $j=30$ grasp attempts (the numbers reported at earlier training episodes {\it i.e.} iteration $i<j$ are weighted by $\frac{i}{j}$). After around 600 training episodes, grasping success rate of goal object becomes stable above 80$\%$ while grasp Q-value stays stably above $Q_g^*=1.8$. Thus it is reasonable to regard $Q_g^*=1.8$ as a threshold to evaluate whether is suitable for goal grasping. 

% {In our system, $Q_g^*$ serves as an estimated value for a successful grasp, which guides pushes to interact with the environment and achieve a graspable state as soon as possible. In Stage II, we fix the grasp net with a fixed threshold $Q_g^*$ to achieve a more stable training. However, there remains the problem of distribution mismatch, thus we add an alternating training stage. In this stage, the threshold $Q_g^*$ should be adjusted indeed since the grasp net is also updated. However, we record the Q value curve of successful grasps (Fig. \ref{fig:q_alternating_training}) during alternating training and find that the Q value for a successful grasp are still stable around 1.8. Therefore, finally we choose a fixed Q threshold in our system.}

{\bf Ablation Studies.} We compare our methods with several ablation methods to test 1) whether our techniques can improve sample efficiency and 2) whether alternating-training can relieve the the problem of distribution mismatch.

We record goal grasp success rate versus training episodes over the last 30 grasp attempts to indicate performance during the push-grasping training process and make comparisons with two ablation methods: {our system without dense reward} is analogous to our method but without any reward for pushing and without any goal-conditioned mechanism, while {our system without goal condition} is a method without any goal-conditioned mechanism but with rewards for pushing.

\begin{figure}[t]
  \centering
  \includegraphics[width=0.8\linewidth]{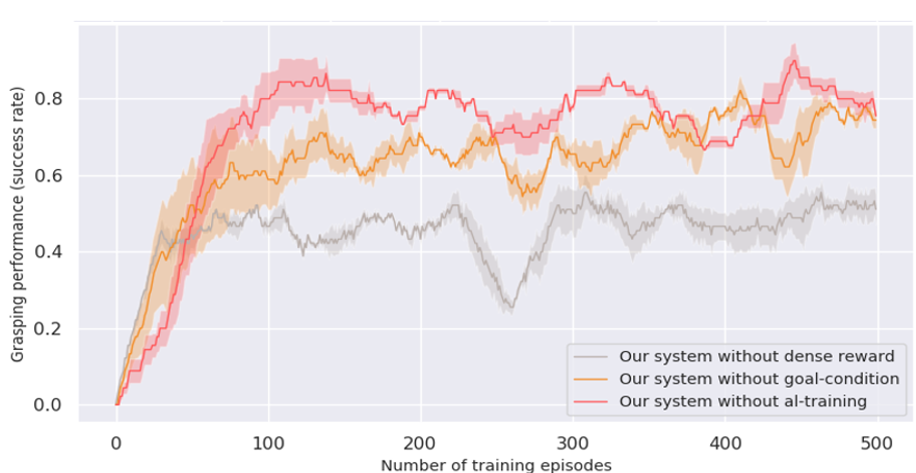}
  \vspace{-0.2cm}
  \caption{Comparing training performance of our method with two ablation methods of our system.}
  \label{fig:ablation}
  \vspace{-0.4cm}
\end{figure}
\begin{figure}[t]
  \centering
  \includegraphics[width=0.8\linewidth]{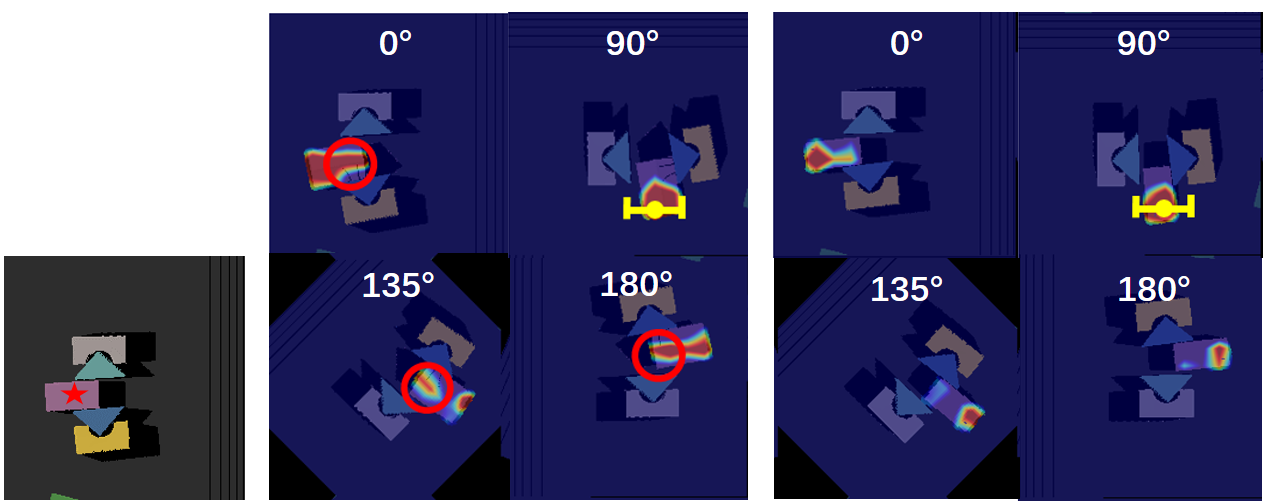}
  \vspace{-0.2cm}
  \caption{Example distributions of grasp Q values within goal object (the purple block) before (middle) and after (right) alternating training. {We choose four typical grasp Q maps corresponding to grasp orientation 0\degree, 90\degree, 135\degree and 180\degree respectively to reflect the effect. The full version can be seen in Appendix-\ref{app-f}}. Color tends to be red as the Q value increasing. Before alternating training, some unreasonable grasps (highlighted with red circles) show high q values. Such low quality grasps are greatly reduced after alternating training. The chosen grasp is represented by a jaw orientation and a middle point of the gripper. }
  \label{fig:alternating_training}
  \vspace{-0.8cm}
\end{figure}
To be rigorous, all methods are trained three times and results are plotted in Fig. \ref{fig:ablation}. {Note that in this figure performance of our system only involves the first two training stages since alternating training is started after 1500 training episodes ({\it {i.e.}} our system without alternating training)}. The curves in the figure show that our system without alternating training is capable of converging to 80$\%$ grasp success rate in just 350 training episodes, which shows the highest training efficiency. This suggests that our adversarial training and goal-conditioned mechanism can effectively improve the sample efficiency so as to speed up the training process. We can see that policy of {our system without dense reward} can keep a similar pace as that of our system without alternating training at the early training but slows down after around 50 episodes. This may be because at early training the pushing policy is raw and does not play an important role for grasping, but after several training episodes, pushing policy with dense rewards has been trained to the extent of easing some grasps while that without dense rewards is still immature. With dense pushing rewards, policy of {our system without goal condition} can achieve a faster learning process and finally reach a similar performance with policy of our system without alternating training in 500 episodes, but still fails in the comparison of the training efficiency because of the lower sample efficiency.

Also we report a case to qualitatively demonstrate the effect of alternating training (more quantitative results are shown in the comparisons to the baseline methods) in Fig. \ref{fig:alternating_training}, which shows example distributions of grasp Q values within goal object (the purple block) before and after alternating training. Obviously, grasping the goal block at positions between the two triangular objects is unreasonable, but it shows a high Q value before alternating training (highlighted with red circles in Fig. \ref{fig:alternating_training}). By means of alternating training, such low quality grasps are removed. Moreover, the grasp chosen before alternating training is at the edge of the goal block which might lead to a fail catching. In contrast, the grasp chosen after alternating training is more stable.

\begin{figure}[t]
  \centering
  \includegraphics[width=0.8\linewidth]{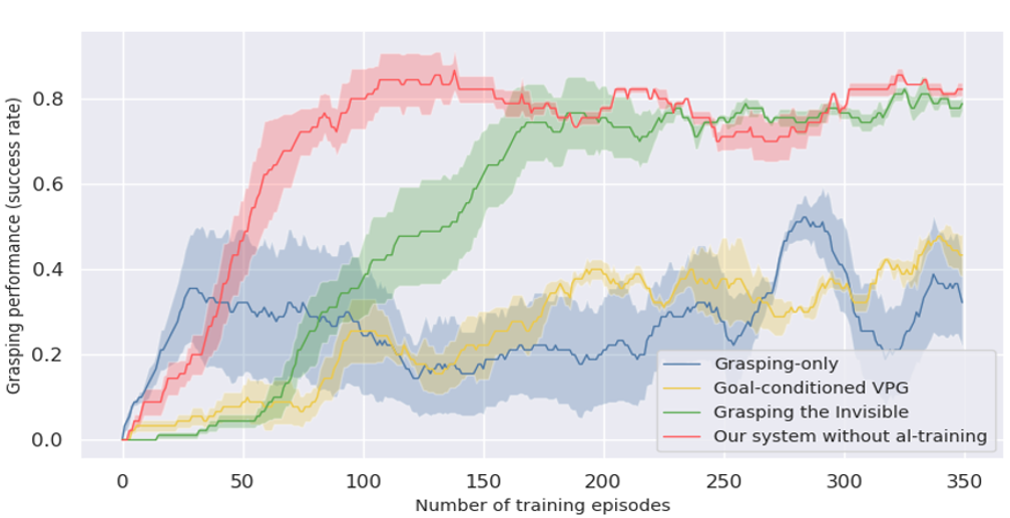}
  \vspace{-0.2cm}
  \caption{Comparing training performance of our method with baselines.}
  \label{fig:baselines}
  \vspace{-0.35cm}
\end{figure}

{\bf Comparisons to baselines.} In the following experiments we compare our system with baselines both in training and testing performances.
Fig \ref{fig:baselines} shows training performance of our methods and baselines. Compared with policy of {\bf Grasping the Invisible}, our system without alternating training reaches 80$\%$ grasp success rate in less training episodes. Performance of {\bf Goal-conditioned VPG} policy is capable of increasing steadily but at a slower pace. As for the {\bf Grasping-only} policy, interestingly it improves its grasping performance at the fastest pace early in training but ultimately achieves the lowest average performance. This may be due to the fact that while the grasping-only policy is busy refining itself to detect harder grasps, others focus on learning pushes that can make grasping easier.

We conduct test experiments with both random and challenging arrangements. For random arrangements, 30 objects are randomly dropped into the workspace whose shapes and colors are randomly chosen during each run. This scene is similar to that of training, but contains 30 objects rather than 10 so as to demonstrate the generalization of policies to more cluttered scenes. We run each test $n = 30$ runs and results are reported in Table \ref{table:1}. We observe that our system has much better performance in task completion rate and grasp success rate using a bit more motion number. It is interesting to note that {\bf Grasping the Invisible} achieves a low grasp success rate at 50-60$\%$. This is likely due to its tendency to grasp when the edge of the goal object is unoccupied but the free space is not enough for the gripper insertion. Besides, although {\bf Goal-condtioned VPG} achieves the least motion number, it does not mean that it has a high efficiency since its grasp success rate is low at 60-70$\%$.

\begin{table}[t]
\caption{{SIMULATION RESULTS ON ALL ARRANGEMENTS}}
\label{table:1}
\vspace{-0.6cm}
\begin{center}
% \begin{tabular}{p{2.2cm}|p{0.25cm}|p{0.25cm}|p{0.25cm}|p{0.25cm}|p{0.25cm}|p{0.25cm}|p{0.25cm}|p{0.25cm}|p{0.25cm}}
\resizebox{\linewidth}{10mm}{
\scriptsize
\begin{tabular}{c|c|c|c|c|c|c}
\hline
\makecell[c]{Method} & \multicolumn{2}{c|}{Completion} & \multicolumn{2}{c|}{Grasp Success} & \multicolumn{2}{c}{Motion Number} \\
\hline
\makecell[c]{Arrangement} & \makecell[c]{r} & \makecell[c]{c} & \makecell[c]{r} & \makecell[c]{c} & \makecell[c]{r} & \makecell[c]{c}\\
\hline
\makecell[c]{Grasping-only} & {90.0} & {78.8} & {60.3} & {29.0} & {4.20} & {7.48}\\
\makecell[c]{Goal-conditioned VPG} & {93.8} & {89.3} & {62.3} & {41.7} & {{\bf 3.93}} & {5.78}\\
\makecell[c]{Grasping the Invisible} & {96.7} & {95.0} & {54.6} & {70.4} & {5.37} & {4.33}\\
{Our system w/o al-training} & {{\bf 97.8}} & {97.7} & {83.7} & {87.6} & {4.50} & {3.20}\\
\makecell[c]{Our system} & {{\bf 97.8}} & {{\bf 99.0}} & {{\bf 90.0}} & {{\bf 90.0}} & {{4.82}} & {{\bf 2.77}}\\
\hline
\end{tabular}}
\begin{tablenotes}
% \tiny
\scriptsize
\item * The second row of table represent arrangement type. r, c correspond to random and challenging arrangements respectively. 
\item ** Detailed tables with metrics of each case are available in Appendix-\ref{app-f}.
\end{tablenotes}
\end{center}
\vspace{-1cm}
\end{table}

\begin{figure}[t]
  \centering
  \includegraphics[width=0.88\linewidth]{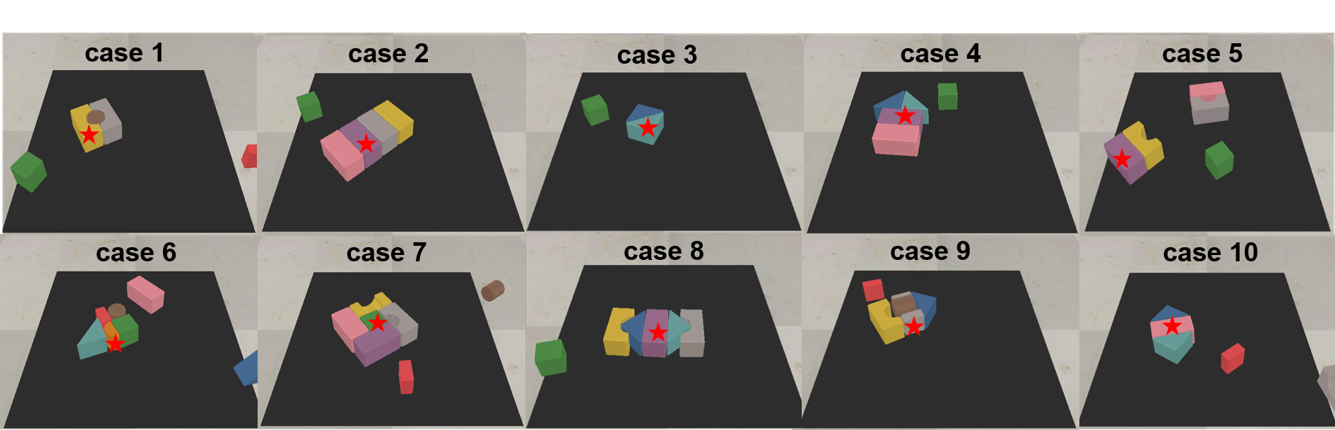}
  \vspace{-0.2cm}
  \caption{Challenging arrangements in simulation experiments which involve 10 designed scenes with adversarial clutter. The goal object of each scenario is labeled with a star.}
  \label{fig:case}
  \vspace{-0.45cm}
\end{figure}

Challenging arrangements involves 10 designed cases (shown in Fig. \ref{fig:case}) with adversarial clutter where pushes are essential for goal grasping since the goal object are arranged either close side by side with others, or encircled by clutter, or even at the edge of the workspace, and it is rather difficult to pick the goal object without de-cluttering. Compared results to all baseline methods after $n=30$ runs can be seen in Table \ref{table:1}.

It is shown in the table that our method outperforms all baselines across all metrics. {\bf Grasping-only} policy achieves the lowest completion rate, and  even when it can complete the task, its average grasp success rate remains low at 20-30$\%$. For push-assisted methods, each of them can achieve a much better completion rate, which suggests that push actions effectively assist goal object grasping in
adversarial clutter. However, goal grasp success rate of {\bf Goal-conditioned VPG} policy remains low at 40-50$\%$, especially in scenes where objects are arranged closely side by side. Although {\bf Grasping the Invisible} policy performs better in grasp success rate than {\bf Goal-conditioned VPG}, it tends to grasp when space around goal objects is still {narrow}, thus its grasp success rate performance is also unsatisfactory. {\bf Grasping the Invisible} policy achieves the lowest grasp success rate and highest number of motion in case 5. This is due to the fact that in this case it tries to grasp the object directly without pushing. And after several continuous failed grasps it tends to push the surroundings but unfortunately, push actions are not succinct and make little difference to the scene.

By training pushing policy with direct supervision of grasp success probability, our policy can perform better in both grasp success rate and motion number. Without alternating training, our policy can achieve an average grasp success rate of 87.6$\%$ (with above 90$\%$ success rate for 5 of 10 test cases) and an average motion number of 3.20, which suggests our push actions are efficient for goal grasping. After alternating training, our policy achieves a higher grasp success rate of 90.2$\%$ and a lower motion number of 2.77, which is apparent in case 6, case 8 and case 10. In case 6, the policy with alternating training can predict grasps at the plane while policy without alternating training tends to grasp at the curved surface.

\begin{figure}[t]
  \centering
  \includegraphics[width=0.68\linewidth]{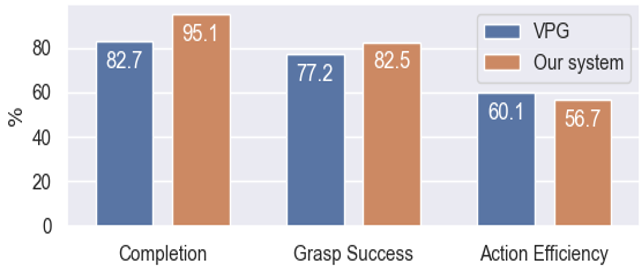}
  \vspace{-0.2cm}
  \caption{Performance of goal-agnostic conditions including completion (left), grasp success rate (middle) and average motion number (right).}
  \label{fig:goal_agnostic_performance}
  \vspace{-0.7cm}
\end{figure}

{\bf Goal-agnostic Conditions.} We further test whether our approach can extend to goal-agnostic conditions. We evaluate our method on 10 challenging cases in Fig. \ref{fig:case} without setting a goal. In this experiment, we mask all objects in the goal mask heightmap and the policy chooses to grasp the one with the highest grasp q value, finally picking up all objects. Results shown in Fig. \ref{fig:goal_agnostic_performance} indicate that our policy can perform well in goal-agnostic conditions too. Compared with {\bf VPG}, our method outperforms in completion rate by 12.4$\%$ and grasp success rate by 5.3$\%$ while action efficiency (defined as $\frac{{\text objects\;in\;test}}{{\text actions\;before\;completion}}$) seems a bit lower. { This might be due to the fact that our policy chooses an object to grasp itself, and seldom turns to grasp others until this object is successfully caught, thus it tends to execute more push actions to create a suitable state for grasping. }  
\vspace{-0.15cm}
\subsection{Real-World Experiments}

\begin{figure}[t]
  \centering
  \includegraphics[width=0.78\linewidth]{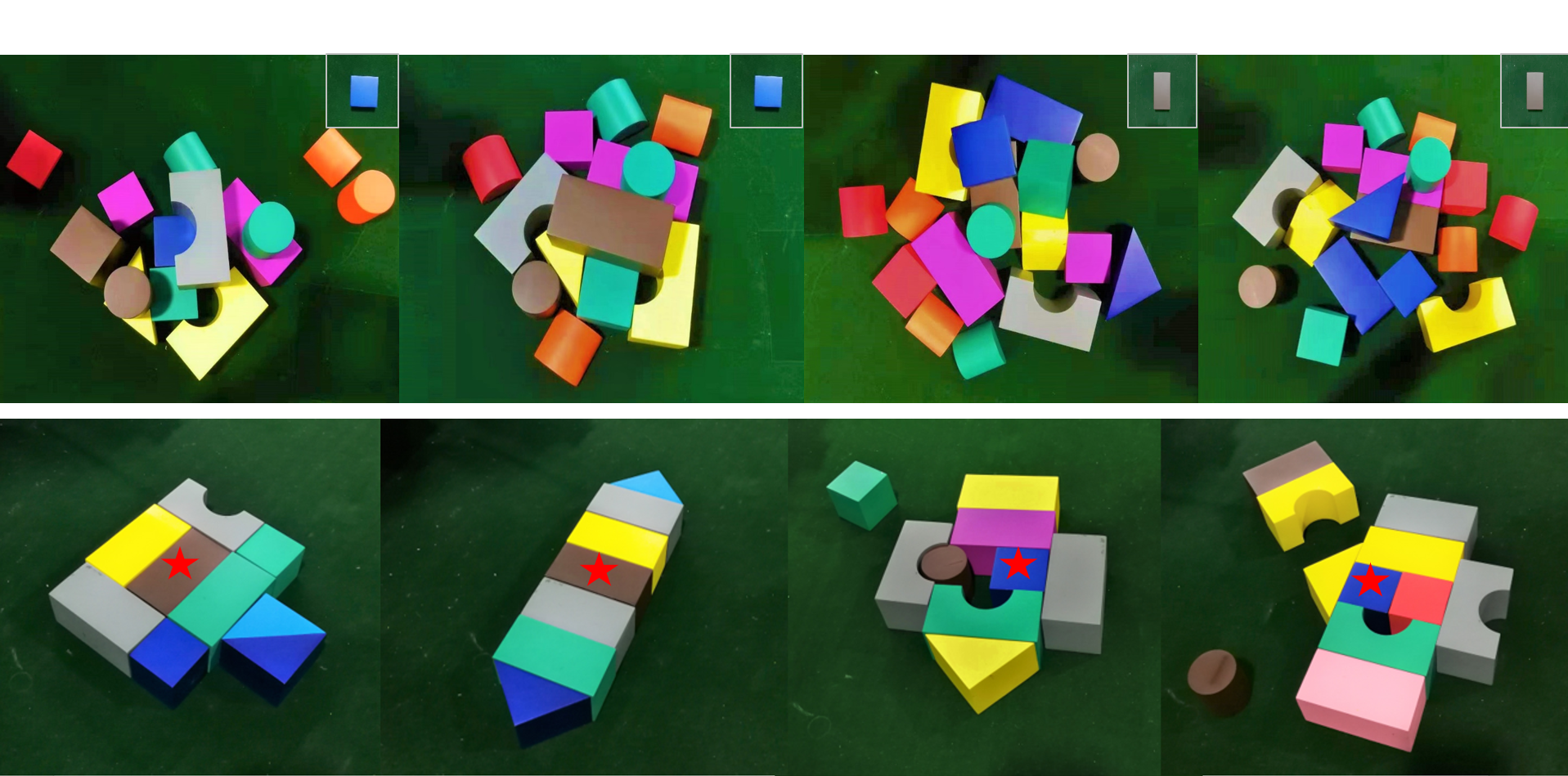}
  \vspace{-0.2cm}
  \caption{{Testing arrangements in real-world experiments. The upper row shows 4 random arrangements in which the goal object shown on the top right corner is partially or fully occluded. Note that the goal object in the second case ({\it i.e.} the blue cube) is fully occluded by the brown block. The lower row shows 4 setting challenging arrangements with each goal object labeled with a star.}}
  \label{fig:real_arrangements}
  \vspace{-0.3cm}
\end{figure}

In this section, we test our system in real-world experiments. Our real-world platform includes a UR5 robot arm with a ROBOTIQ-85 gripper. RGB-D images of resolution $1280 \times 720$ are captured from an Intel RealSense D435. {Our test cases consist of 4 random arrangements in which the goal object is partially or fully occluded and 4 setting challenging arrangements (shown in Fig. \ref{fig:real_arrangements}).} We compare our methods with the method {\bf Grasping the Invisible} which has shown a better performance than other baselines in our simulation experiments and other related works in \cite{yang2020deep}.

In the real-world experiments, we set $n=15$ runs for each test. Note, models of both methods are transferred from simulation to the real world without any retraining. Details about task completion rate, grasp success rate and average motion number of each case are reported in Table \ref{table:3}. Overall, our policy outperforms the baseline {across all metrics in random and challenging settings}. This suggests that our policy can effectively generalize to the real world setting. 
% {\bf Grasping the Invisible} {achieves} a lower completion rate by 10$\%$ and a lower grasp success rate by 16.3$\%$. 
It is worth noting that {\bf Grasping the Invisible} uses a classifier to decide the actions, which largely depends on the accuracy of the depth images. This might partly explain why their policy is usually unable to accurately judge when to grasp.

\begin{figure}[t]
  \centering
  \includegraphics[width=0.78\linewidth]{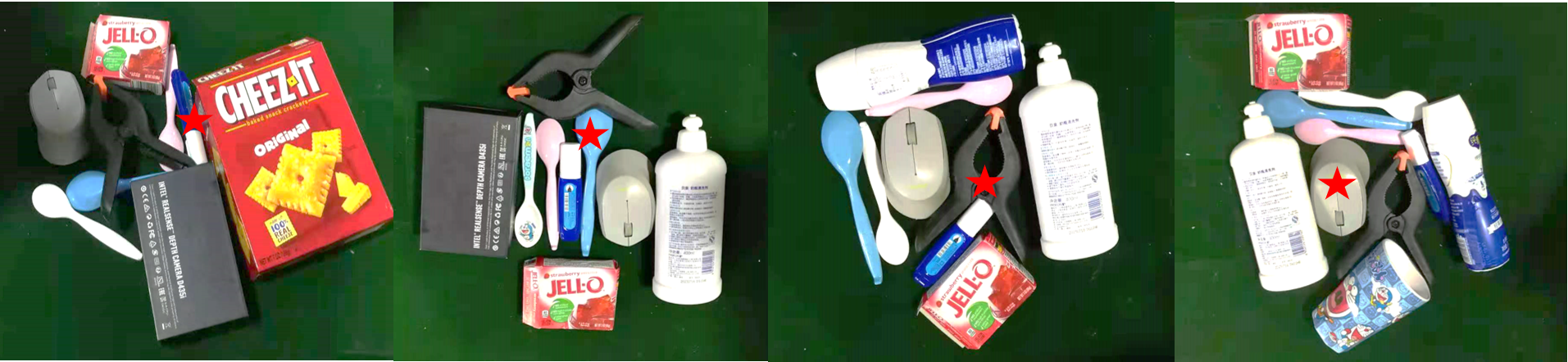}
  \vspace{-0.2cm}
  \caption{{Real-world testing cases containing arrangements with novel objects which is unseen during training process.}}
  \label{fig:real_novel}
  \vspace{-0.7cm}
\end{figure}

{Furthermore, we test the generalization capability of novel objects unseen during training. Each scenario contains objects of different height and more complex shape (shown in Fig. \ref{fig:real_novel}). Results in Table \ref{table:3} confirm that our method can generalize to some unseen objects with a similar object shape distribution as that of the training objects, and has better performance than other methods. Failed cases involve catching an object with completely unseen complex shape, which might due to the lack of grasping experience of such kind of shape during training process.}

\begin{table}[t]
\caption{{REAL-WORLD RESULTS ON ALL ARRANGEMENTS}}
\label{table:3}
\vspace{-0.6cm}
\begin{center}
% \begin{tabular}{p{2.2cm}|p{0.25cm}|p{0.25cm}|p{0.25cm}|p{0.25cm}|p{0.25cm}|p{0.25cm}|p{0.25cm}|p{0.25cm}|p{0.25cm}}
\resizebox{1.03\linewidth}{6mm}{
\scriptsize
\begin{tabular}{c|c|c|c|c|c|c|c|c|c}
\hline
\makecell[c]{Method} & \multicolumn{3}{c|}{Completion} & \multicolumn{3}{c|}{Grasp Success} & \multicolumn{3}{c}{Motion Number} \\
\hline
\makecell[c]{Arrangement} & \makecell[c]{r} & \makecell[c]{c} & \makecell[c]{n}  & \makecell[c]{r} & \makecell[c]{c} & \makecell[c]{n} & \makecell[c]{r} & \makecell[c]{c} & \makecell[c]{n}\\
\hline
{Grasping the Invisible} & {86.7} & {85.0} & {88.3} & {75.2} & {70.3} & {65.1} & {6.92} & {6.81} & {{\bf 3.81}}\\
\makecell[c]{Our system} & {{\bf 93.3}} & {{\bf 95.0}} & {{\bf 90.0}} & {{\bf 81.7}} & {{\bf 86.6}} & {{\bf 81.0}} & {{\bf 5.67}} & {{\bf 4.62}} & {4.45}\\
\hline
\end{tabular}}
\begin{tablenotes}
% \tiny
\scriptsize
\item * The second row of table represent arrangement type. r, c, n correspond to random, challenging and novel object arrangements respectively. 
\item ** Detailed tables with metrics of each case are available in Appendix-\ref{app-f}.
\end{tablenotes}
\end{center}
\vspace{-0.9cm}
\end{table}

% \begin{table}[h]
% \caption{{REAL-WORLD RESULTS ON CHALLENGING ARRANGEMENTS}}
% \label{table:4}
% \vspace{-0.3cm}
% \scriptsize
% \begin{center}
% \begin{tabular}{p{2.7cm}|p{1.05cm}|p{1.5cm}|p{1.65cm}}
% \hline
% \makecell[c]{Method} & Completion & \makecell[c]{Grasp Success} & \makecell[c]{Motion Number} \\
% \hline
% \makecell[c]{Grasping the Invisible} & \makecell[c]{85.0} & \makecell[c]{70.3} & \makecell[c]{6.81}\\
% \makecell[c]{Our system} & \makecell[c]{{\bf 95.0}} & \makecell[c]{{\bf 86.6}} & \makecell[c]{{\bf 4.62}}\\
% \hline
% \end{tabular}
% \end{center}
% \vspace{-0.5cm}
% \end{table}

% \begin{table}[h]
% \caption{{REAL-WORLD RESULTS ON NOVEL OBJECT ARRANGEMENTS}}
% \label{table:real-novel}
% \vspace{-0.3cm}
% \scriptsize
% \begin{center}
% \begin{tabular}{p{2.7cm}|p{1.05cm}|p{1.5cm}|p{1.65cm}}
% \hline
% \makecell[c]{Method} & Completion & \makecell[c]{Grasp Success} & \makecell[c]{Motion Number} \\
% \hline
% \makecell[c]{Grasping the Invisible} & \makecell[c]{88.3} & \makecell[c]{65.1} & \makecell[c]{{\bf 3.81}}\\
% \makecell[c]{Our system} & \makecell[c]{{\bf 90.0}} & \makecell[c]{{\bf 81.0}} & \makecell[c]{{4.45}}\\
% \hline
% \end{tabular}
% \end{center}
% \vspace{-0.8cm}
% \end{table}

%%%%%%%%%%%%%%%%%%%%%%%%%%%%%%%%%%%%%%%%%%%%%%%%%%%%%%%%%%%%%%%%%%%%%%%%%%%%%%%%
% \vspace{-0.1cm}
\section{CONCLUSION}
% \vspace{-0.1cm}
In this work, we study the task of grasping a goal object in clutter, which is formulated as a goal-conditioned hierarchical reinforcement learning problem. We train the policies in simulation with three stages and evaluate our system in both simulation and real-world platforms. Simulation experiments show that our system can speed up the training process compared with other methods and achieve better performance in both random and challenging cases. Real-world experiments suggest our system is capable of effectively generalizing from simulation to the real world. It is worth mentioning that our system also achieves a better performance in goal-agnostic conditions, which further extends application scenarios of our method. {More analysis and experiments can be accessed at Appendix}.
\vspace{-0.3cm}

%%%%%%%%%%%%%%%%%%%%%%%%%%%%%%%%%%%%%%%%%%%%%%%%%%%%%%%%%%%%%%%%%%%%%%%%%%%%%%%%

\ifCLASSOPTIONcaptionsoff
  \newpage
\fi



%\bibliographystyle{IEEEtran}

\bibliographystyle{IEEEtran}
\bibliography{IEEEabrv,ref}

\appendix
\label{appendix}
\subsection{More analysis of the threshold $Q_g^*$}
\label{app-a}
In our system, $Q^*$ serves as an estimated value for a successful grasp. With this threshold, we can encourage the early ending of an episode if pushes have created a graspable scene, which guides the robot to execute push actions as few as possible. This setting makes our learning more simple and efficient. Without this setting, the robot must execute a fixed number of pushes before a grasp, or just compare the Q value of pushing and grasping ({\it i.e.} VPG), or require additional training of a action classifier to choose between pushing and grasping ({\it i.e.} Grasping the Invisible). Lots of qualitative and quantitative results in our paper have also confirmed that our method is more efficient and has a better performance.

Also, we set a fixed value because we think $Q^*$ will not change with the scenario or the length of action sequence, but mostly based on the state of the goal object. In Stage II, we fix the grasp net with a fixed threshold $Q^*$ to training pushes to create a graspable state. In this stage, the grasp net and the threshold $Q^*$ can be regarded as the environment for the pushing policy. Similarly, in alternating training stage, the threshold $Q^*$ is also fixed for learning a strategy to reach a stable graspable state as soon as possible in each episode. This threshold can also be regarded as a part of environment. Compared with other methods which uses a graspable probability through softmax, this setting may be more strict.  

Besides, $Q^*$ depends the graspability of the goal object we want to achieve. That is, if we set $Q^*=2.0$, then finally Q value of successful grasps will converge to 2.0. But more push actions will be needed to build this system.

\subsection{Completely handcrafted vs. data-driven pushing rewards}
\label{app-b}
In this section, we will elaborate why completely handcrafted pushing reward for goal-oriented grasping task is limited and defective, thus confirming the necessity of a data-driven pushing rewards.

First, we agree that we can completely handcraft a reward function if our intuition is already good on the problem. But unfortunately, our intuition about how pushes enable future grasps is still not mature enough. Intrinsic pushing reward in VPG \cite{zeng2018learning} does not explicitly consider whether a push enables future grasps. Rather, it simply encourages the system to make pushes that cause change. Also, GTI \cite{yang2020deep} develops denser rewards by assigning rewards for pushing totally with handcraft to encourage moving objects and removing clutter. These rewards are designed from human intuition that pushes might be supposed to move objects and remove clutter around a goal object. However, it's difficult to quantify the amount of space around the object for grasping. Encouraging pushes to crate space might lead to push unnecessarily for many times without a grasp when it is not confident enough to grasp because the occupancy rate is not low enough. In some cases where the goal object is higher than the surroundings, the object can be directly grasped without any space around in the top-down view. Besides, even the occupancy around the goal object is low, it does not mean that there is enough space for the gripper insertion. In \cite{danielczuk2018linear}, the authors design a series of handcrafted pushing reward functions and conduct studies about how these rewards assist future grasps, which suggests that the graspable probability is not always correlated with object separation.

Also, our pushing reward function is handcrafted in form. However, we use a grasp net to measure the push rewards, thus the value of pushing reward is essentially data-driven. Our insight is to mainly use a pre-trained grasp net to guide the training of the pushing policy, combined with a proxy reward designed from human intuition. The handcrafted part serves as prior knowledge which leverages human intuition to encourage pushes to remove clutter around the goal object, which can effectively speed up the training process. More importantly, we encourage pushes that directly improve the graspable probability of the goal object predicted by the pre-trained grasp net. In this way, we explicitly consider whether a push enables future grasps and utilize posterior experience learned by the grasp net to optimize the system.

\begin{figure}[t]
  \centering
  \includegraphics[width=0.8\linewidth]{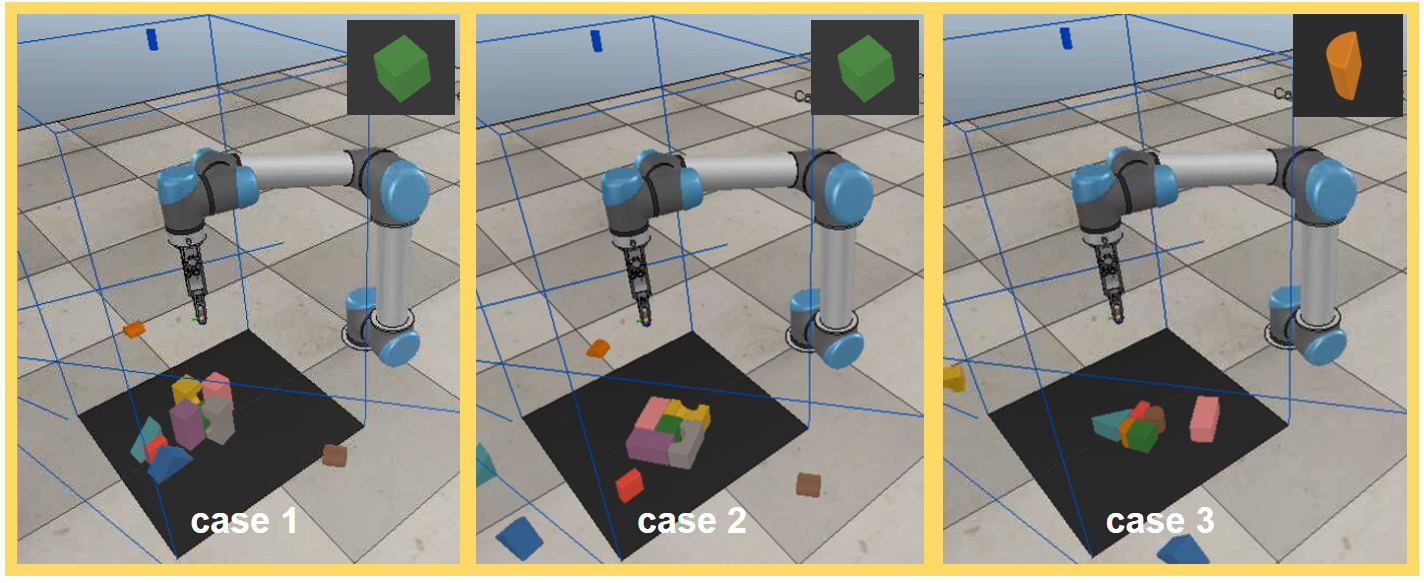}
  \caption{Testing cases to show the weakness of totally handcrafted reward. The goal object of each case is shown on the top right corner.}
  \label{fig:case_study}
%   \vspace{-0.4cm}
\end{figure}

\begin{figure}[t]
\centering
\subfigure[Ours]{
\begin{minipage}[b]{\linewidth}
\includegraphics[width=\linewidth]{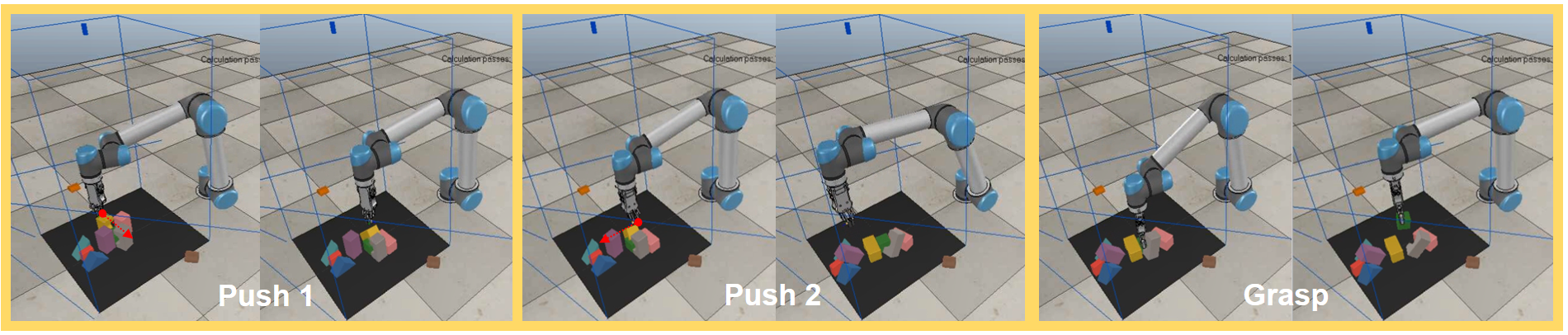}
\end{minipage}
}
\subfigure[Grasping the Invisible]{
\begin{minipage}[b]{\linewidth}
\includegraphics[width=\linewidth]{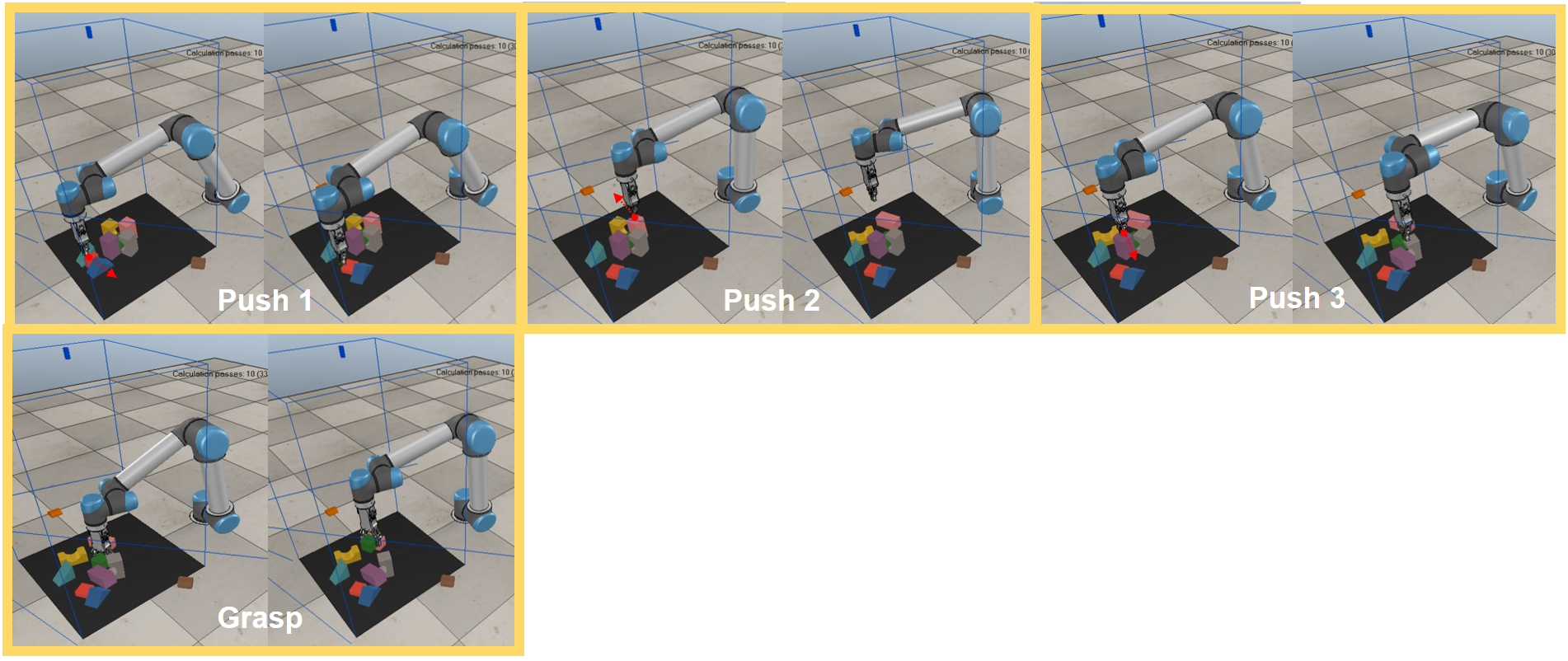}
\end{minipage}
}
\caption{Example action sequences in testing case 1 of two methods.} \label{fig:case_1}
% \vspace{-0.5cm}
\end{figure}

\begin{figure}[t]
\centering
\subfigure[Ours]{
\begin{minipage}[b]{0.67\linewidth}
\includegraphics[width=\linewidth]{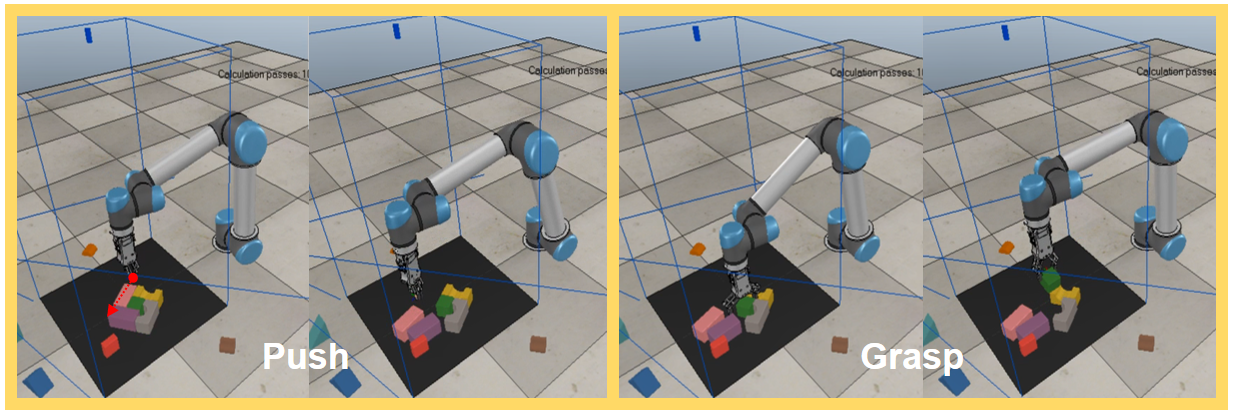}
\end{minipage}
}
\subfigure[Grasping the Invisible]{
\begin{minipage}[b]{\linewidth}
\includegraphics[width=\linewidth]{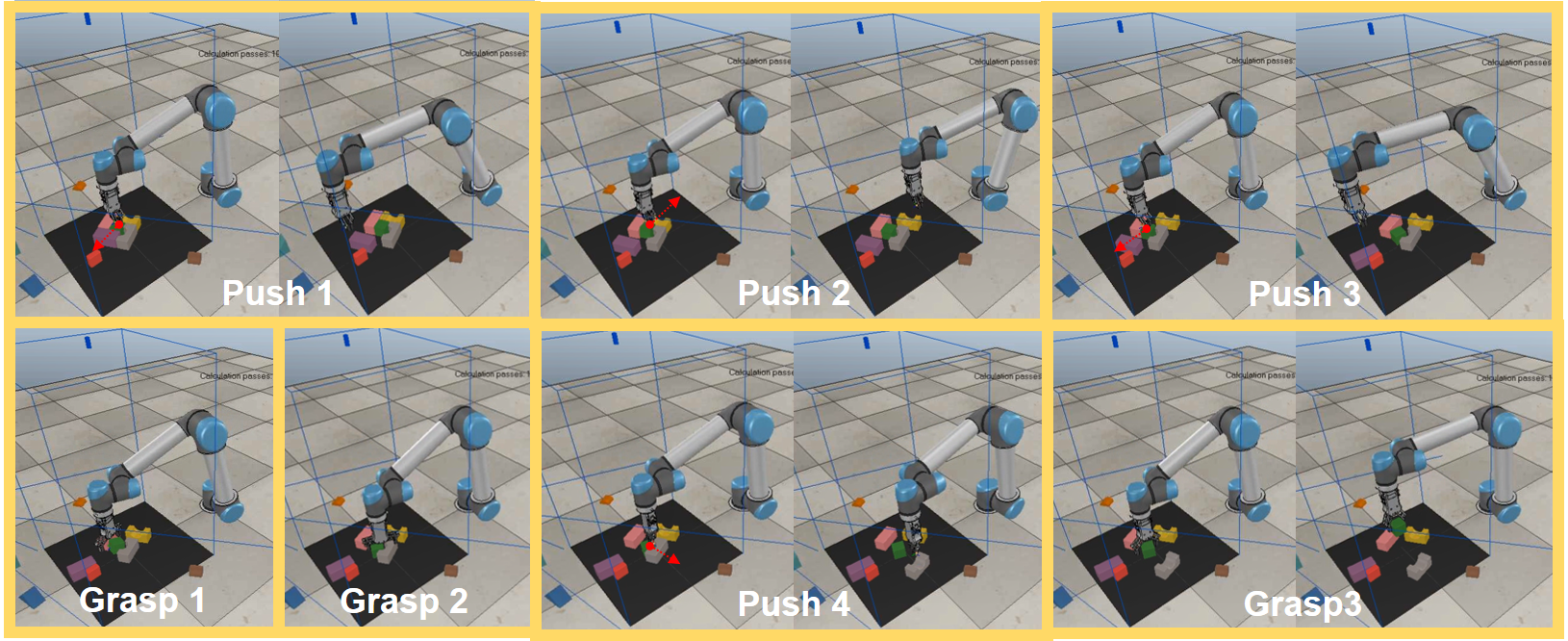}
\end{minipage}
}
\caption{Example action sequences in testing case 2 of two methods.} \label{fig:case_2}
% \vspace{-0.5cm}
\end{figure}

\begin{figure}[!h]
\centering
\subfigure[Ours]{
\begin{minipage}[b]{0.67\linewidth}
\includegraphics[width=\linewidth]{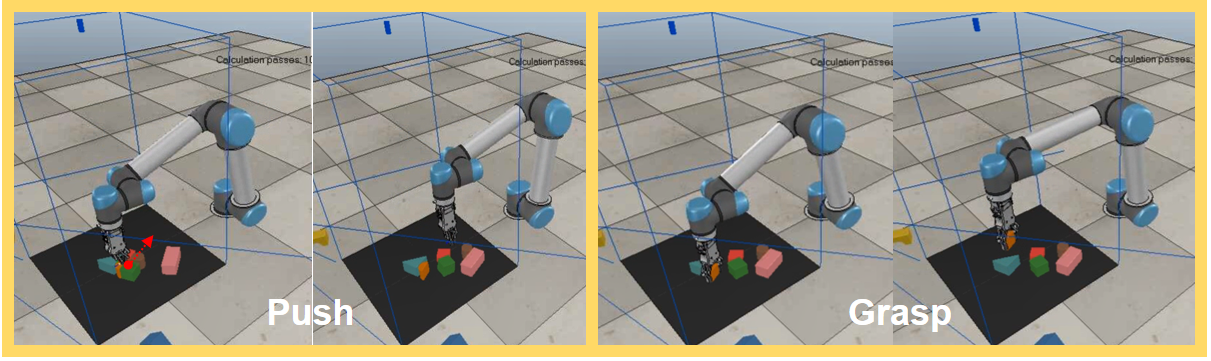}
\end{minipage}
}
\subfigure[Grasping the Invisible]{
\begin{minipage}[b]{0.67\linewidth}
\includegraphics[width=\linewidth]{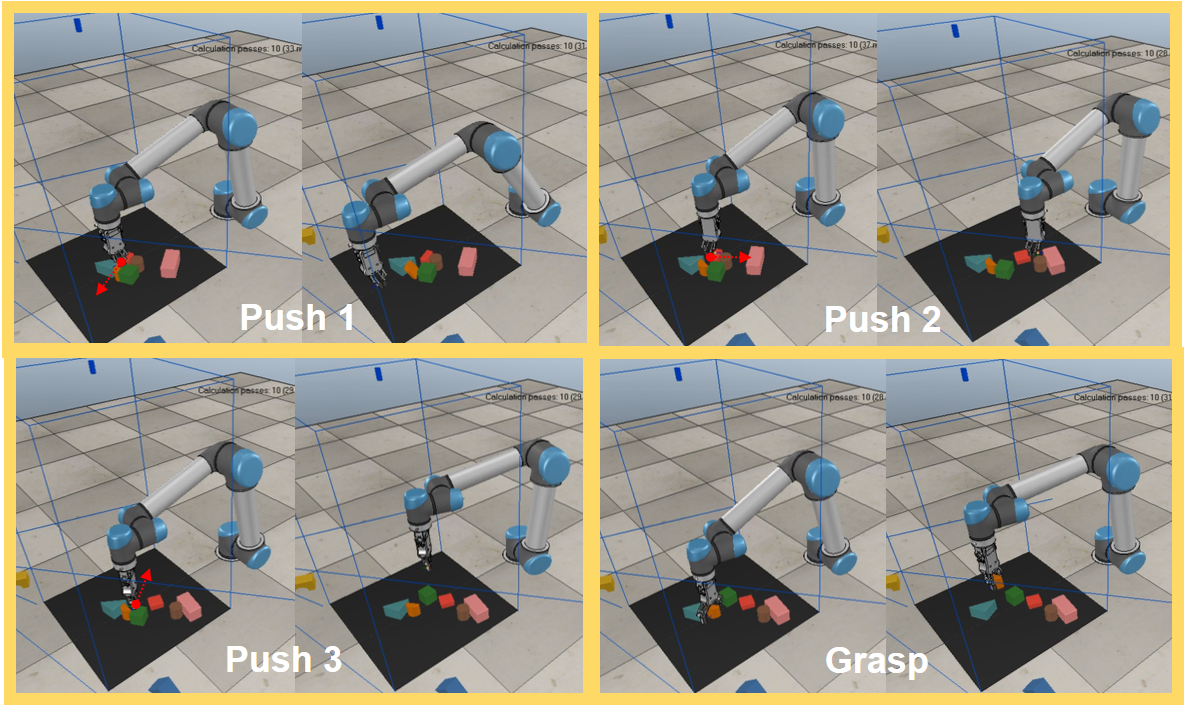}
\end{minipage}
}
\caption{Example action sequences in testing case 3 of two methods.} \label{fig:case_3}
% \vspace{-0.5cm}
\end{figure}

To be more specific, we conduct some case studies to verify the above points. We compare our method with GTI (Grasping the Invisible) \cite{yang2020deep} with 3 testing cases shown in Fig. \ref{fig:case_study}. We can see that GTI, which uses complete handcrafted pushing rewards to encourage pushes to move or create space around the goal object, is always not confident for grasping and tend to push many times to make the surrounding of the goal object free enough (Fig. \ref{fig:case_1}-\ref{fig:case_3}). Also, in case 1, GTI executes a redundant push action (Push 1 in Fig. \ref{fig:case_1}) which changes the environment but is useless for goal object grasping. In contrast, our method can detect effective grasps even if the free space around the goal object is narrow (Fig. \ref{fig:case_1}-\ref{fig:case_2}).

\subsection{Comparison of data requirement}
\label{app-c}
\begin{figure}[!h]
  \centering
  \includegraphics[width=0.88\linewidth]{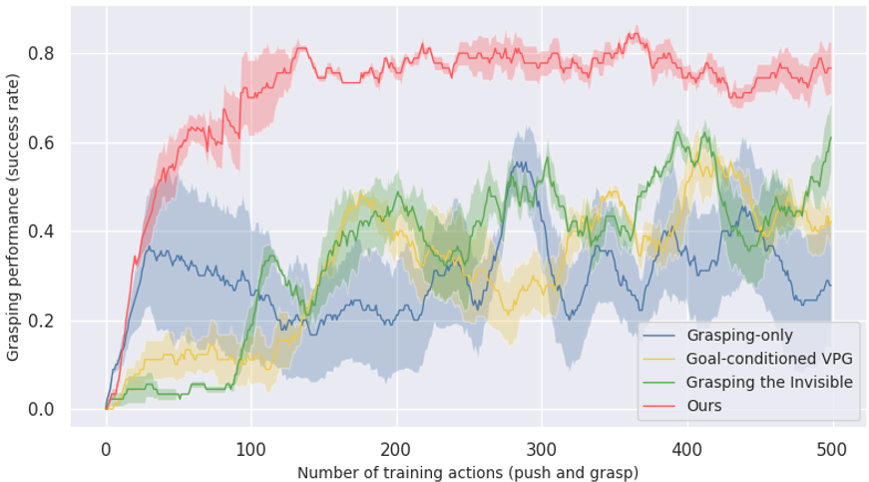}
  \vspace{-0.2cm}
  \caption{{Training curves of goal grasp success rate over the number of training actions (push and grasp).}}
  \label{fig:data_requirement}
%   \vspace{-0.2cm}
\end{figure}

{Also, to measure the data requirement of our method and other baselines, we record goal grasp success rate over the number of training actions in Fig. \ref{fig:data_requirement}. For each action step, the goal grasp success rate is measured by the average goal grasp success rate over the last 30 training actions. Results shows that our method needs the minimum number of training actions (pushes and grasps) to reach 80$\%$ grasping success rate, which means the minimum requirement of data. Besides, combining Fig. \ref{fig:baselines}, we can draw the conclusion that our push actions are more effective than other methods, especially compared to Grasping the Invisible. Although it can converge to a 80$\%$ goal grasp success rate in around 300 grasp attempts, it consumes much more push actions.}

\subsection{Simulated experiments of object generalization}
\label{app-d}
\begin{figure}[t]
  \centering
  \includegraphics[width=\linewidth]{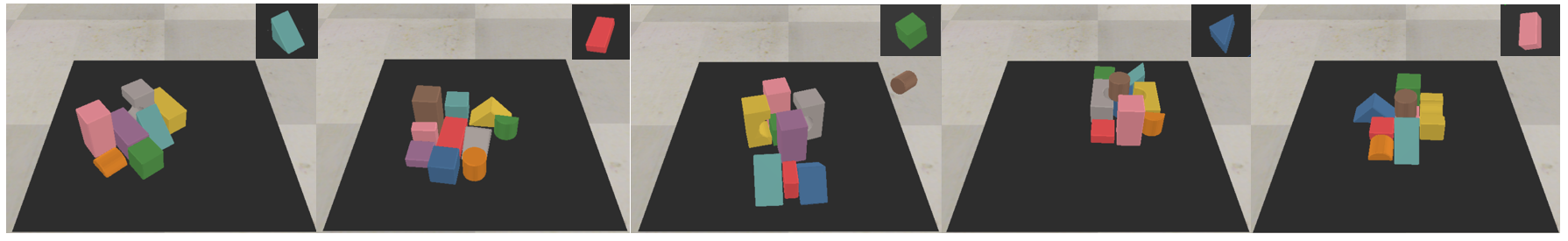}
  \caption{Testing cases containing objects with different heights. The goal object of each scenario is shown on the top right corner.}
  \label{fig:simulated_height}
%   \vspace{-0.6cm}
\end{figure}

{\bf Objects of different heights.} Five additional testing cases (shown in Fig. \ref{fig:simulated_height}) which contain objects with different heights are conducted. The comparison shows the advantage of our method (shown in Table \ref{table:simulated-height}).

\begin{figure}[t]
  \centering
  \includegraphics[width=\linewidth]{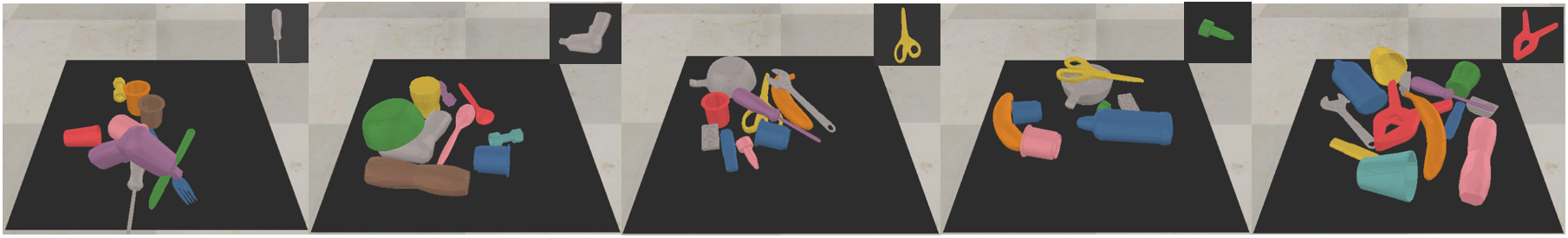}
  \caption{{Simulated testing cases containing arrangements with novel objects which is unseen during training process. Also, objects in these arrangements are of relatively complex shape and different height.}}
  \label{fig:simulated_novel}
%   \vspace{-0.6cm}
\end{figure}

{\bf Objects of complex shape.} Furthermore, we conduct experiments with novel objects of more complex shape than those during training. Cases are shown in Fig. \ref{fig:simulated_novel} and results are reported in Table \ref{table:simulated-novel}. Overall, our method can generalize to some unseen objects with a similar object shape distribution as that of the training objects. Failed cases involve catching an object with completely unseen complex shape, which might due to the lack of grasping experience of such kind of shape during training process.

\begin{table}[t]
\caption{SIMULATION RESULTS ON ARRANGEMENTS WITH DIFFERENT HEIGHT OBJECTS}
\label{table:simulated-height}
\scriptsize
\vspace{-0.3cm}
\begin{center}
\begin{tabular}{p{2.7cm}|p{1.05cm}|p{1.5cm}|p{1.65cm}}
\hline
\makecell[c]{Method} & {Completion} & \makecell[c]{Grasp Success} & \makecell[c]{Motion Number} \\
\hline
\makecell[c]{Goal-conditioned VPG} & \makecell[c]{{\bf 98.7}} & \makecell[c]{59.4} & \makecell[c]{5.58}\\
\makecell[c]{Grasping the Invisible} & \makecell[c]{90.7} & \makecell[c]{23.8} & \makecell[c]{7.89}\\
\makecell[c]{Our system} & \makecell[c]{97.3} & \makecell[c]{{\bf 87.3}} & \makecell[c]{{\bf 5.02}}\\
\hline
\end{tabular}
\end{center}
% \vspace{-0.6cm}
\end{table}

\begin{table}[t]
\caption{{SIMULATION RESULTS ON NOVEL OBJECT ARRANGEMENTS}}
\label{table:simulated-novel}
\vspace{-0.3cm}
\scriptsize
\begin{center}
\begin{tabular}{p{2.7cm}|p{1.05cm}|p{1.5cm}|p{1.65cm}}
\hline
\makecell[c]{Method} & {Completion} & \makecell[c]{Grasp Success} & \makecell[c]{Motion Number} \\
\hline
\makecell[c]{Grasping-only} & \makecell[c]{86.7} & \makecell[c]{44.9} & \makecell[c]{8.14}\\
\makecell[c]{Goal-conditioned VPG} & \makecell[c]{{\bf 100}} & \makecell[c]{60.2} & \makecell[c]{6.63}\\
\makecell[c]{Grasping the Invisible} & \makecell[c]{87.3} & \makecell[c]{57.9} & \makecell[c]{5.81}\\
Our system w/o al-training & \makecell[c]{97.3} & \makecell[c]{73.2} & \makecell[c]{5.85}\\
\makecell[c]{Our system} & \makecell[c]{97.3} & \makecell[c]{{\bf 86.2}} & \makecell[c]{{\bf 4.75}}\\
\hline
\end{tabular}
\end{center}
% \vspace{-0.6cm}
\end{table}

\begin{figure}[!h]
  \centering
  \includegraphics[width=0.7\linewidth]{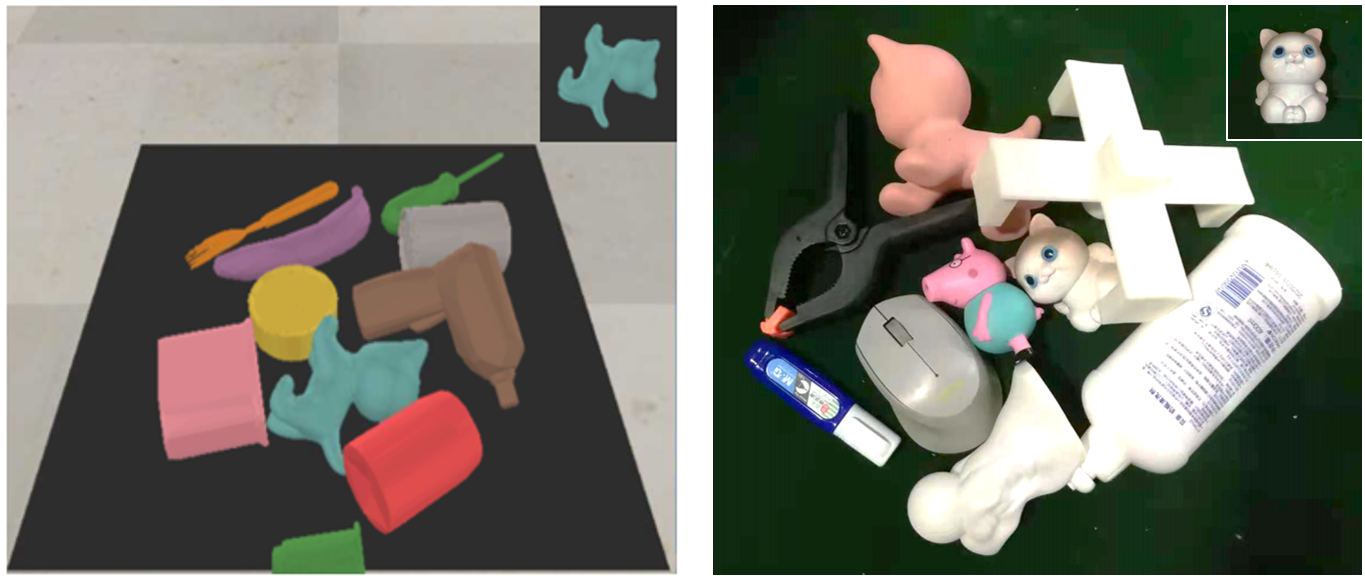}
  \caption{Examples of failed cases. The goal object is a toy cat of unseen complex shape.}
  \label{fig:fail_case}
%   \vspace{-0.4cm}
\end{figure}

{\bf Example failed cases.} Typical failed cases are shown in Fig. \ref{fig:fail_case}), which involve catching an object with unseen complex shape out of the object shape distribution as that of training objects. By applying some advanced object pose estimation methods or enriching the object shape ranges during training process, we think our method can generalize to more complex object shapes.

\subsection{Example action sequence in real-world experiments}
\label{app-e}
\begin{figure}[!h]
  \centering
  \includegraphics[width=\linewidth]{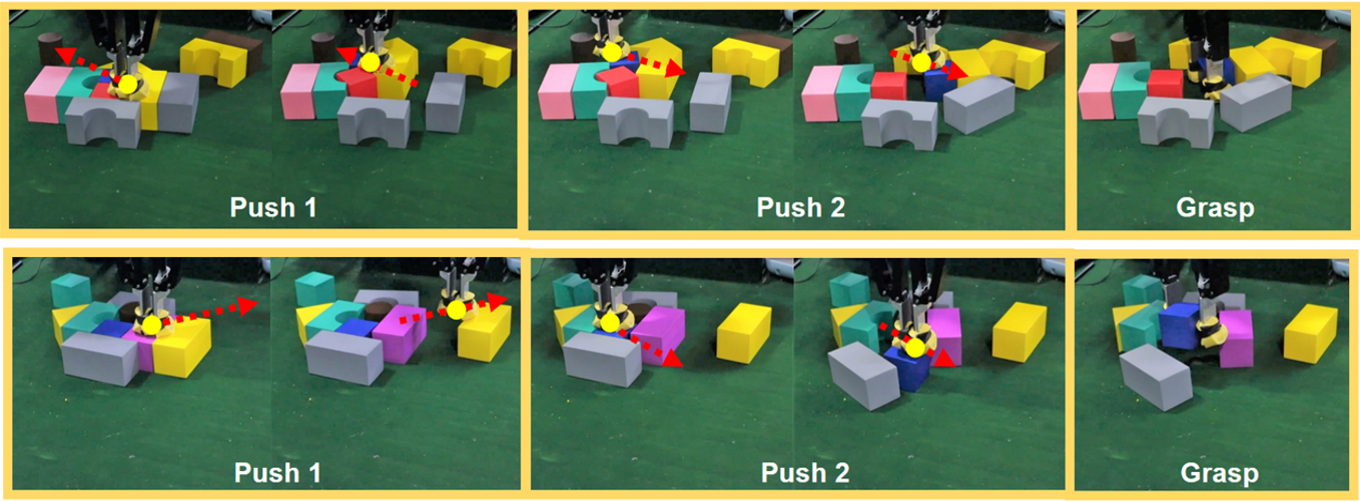}
  \caption{Example action sequences of real-world experiments. The upper row corresponds to the third challenging arrangement while the lower row to the fourth one.}
  \label{fig:real_case}
%   \vspace{-0.5cm}
\end{figure}

Two example action sequences in real-world experiments are shown in Fig. \ref{fig:real_case}.

\begin{table}[t]
% \vspace{0.2cm}
\caption{{REAL-WORLD RESULTS ON RANDOM ARRANGEMENTS}}
\label{table:3-1}
\vspace{-0.3cm}
\begin{center}
\scriptsize
\begin{tabular}{p{1cm}|p{2.7cm}|p{2.7cm}}
\hline
 & \makecell[c]{Grasping the Invisible} & \makecell[c]{Our system} \\
\hline
\makecell[c]{case1} & \makecell[c]{93.3 / 68.8 / 6.88} & \makecell[c]{93.3 / 78.6 / 5.50}\\
\makecell[c]{case2} & \makecell[c]{86.7 / 71.4 / 6.33} & \makecell[c]{100 / 83.3 / 3.87}\\
\makecell[c]{case3} & \makecell[c]{86.7 / 85.7 / 7.14} & \makecell[c]{93.3 / 88.1 / 5.86}\\
\makecell[c]{case4} & \makecell[c]{80.0 / 75.0 / 7.33} & \makecell[c]{86.7 / 76.9 / 7.46}\\
\makecell[c]{average} & \makecell[c]{86.7 / 75.2 / 6.92} & \makecell[c]{{\bf 93.3} / {\bf 81.7} / {\bf 5.67}}\\
\hline
\end{tabular}
\begin{tablenotes}
\scriptsize
     \item * Number of each cell is presented as “Completion / Grasp Success / Motion Number” to display the detailed testing performance of all the cases in the upper row of Figure \ref{fig:real_arrangements}.
\end{tablenotes}
\end{center}
\vspace{-0.4cm}
\end{table}

\begin{table}[t]
% \vspace{0.2cm}
\caption{REAL-WORLD RESULTS ON CHALLENGING ARRANGEMENTS}
\label{table:4-1}
\vspace{-0.3cm}
\begin{center}
\scriptsize
\begin{tabular}{p{1cm}|p{2.7cm}|p{2.7cm}}
\hline
 & \makecell[c]{Grasping the Invisible} & \makecell[c]{Our system} \\
\hline
\makecell[c]{case1} & \makecell[c]{86.7 / 73.3 / 6.75} & \makecell[c]{93.3 / 85.7 / 4.78}\\
\makecell[c]{case2} & \makecell[c]{80.0 / 70.4 / 6.83} & \makecell[c]{100 / 90.0 / 3.40}\\
\makecell[c]{case3} & \makecell[c]{80.0 / 64.8 / 7.11} & \makecell[c]{100 / 90.0 / 5.33}\\
\makecell[c]{case4} & \makecell[c]{93.3 / 72.7 / 6.54} & \makecell[c]{86.7 / 80.8 / 4.97}\\
\makecell[c]{average} & \makecell[c]{85.0 / 70.3 / 6.81} & \makecell[c]{{\bf 95.0} / {\bf 86.6} / {\bf 4.62}}\\
\hline
\end{tabular}
\end{center}
\begin{tablenotes}
\scriptsize
     \item * Number of each cell is presented as “Completion / Grasp Success / Motion Number” to display the detailed testing performance of all the cases in the lower row of Figure \ref{fig:real_arrangements}.
\end{tablenotes}
\vspace{-0.6cm}
\end{table}

\begin{table}[t]
% \vspace{0.2cm}
\caption{REAL-WORLD RESULTS ON NOVEL OBJECT ARRANGEMENTS}
\label{table:5-1}
\vspace{-0.3cm}
\begin{center}
\scriptsize
\begin{tabular}{p{1cm}|p{2.7cm}|p{2.7cm}}
\hline
 & \makecell[c]{Grasping the Invisible} & \makecell[c]{Our system} \\
\hline
\makecell[c]{case1} & \makecell[c]{86.7 / 75.0 / 4.52} & \makecell[c]{93.3 / 86.7 / 4.52}\\
\makecell[c]{case2} & \makecell[c]{93.3 / 80.1 / 3.87} & \makecell[c]{86.7 / 80.6 / 5.54}\\
\makecell[c]{case3} & \makecell[c]{93.3 / 50.0 / 2.00} & \makecell[c]{86.7 / 79.2 / 4.08}\\
\makecell[c]{case4} & \makecell[c]{80.0 / 55.1 / 4.83} & \makecell[c]{93.3 / 77.4 / 3.64}\\
\makecell[c]{average} & \makecell[c]{88.3 / 65.1 / {\bf 3.81}} & \makecell[c]{{\bf 90.0} / {\bf 81.0} / {4.45}}\\
\hline
\end{tabular}
\end{center}
\begin{tablenotes}
\scriptsize
     \item * Number of each cell is presented as “Completion / Grasp Success / Motion Number” to display the detailed testing performance of all the cases in the lower row of Figure \ref{fig:real_novel}.
\end{tablenotes}
\vspace{-0.6cm}
\end{table}

\subsection{Detailed tables and clearer figures}
\label{app-f}
\begin{figure*}[b]
  \centering
  \includegraphics[width=0.88\textwidth]{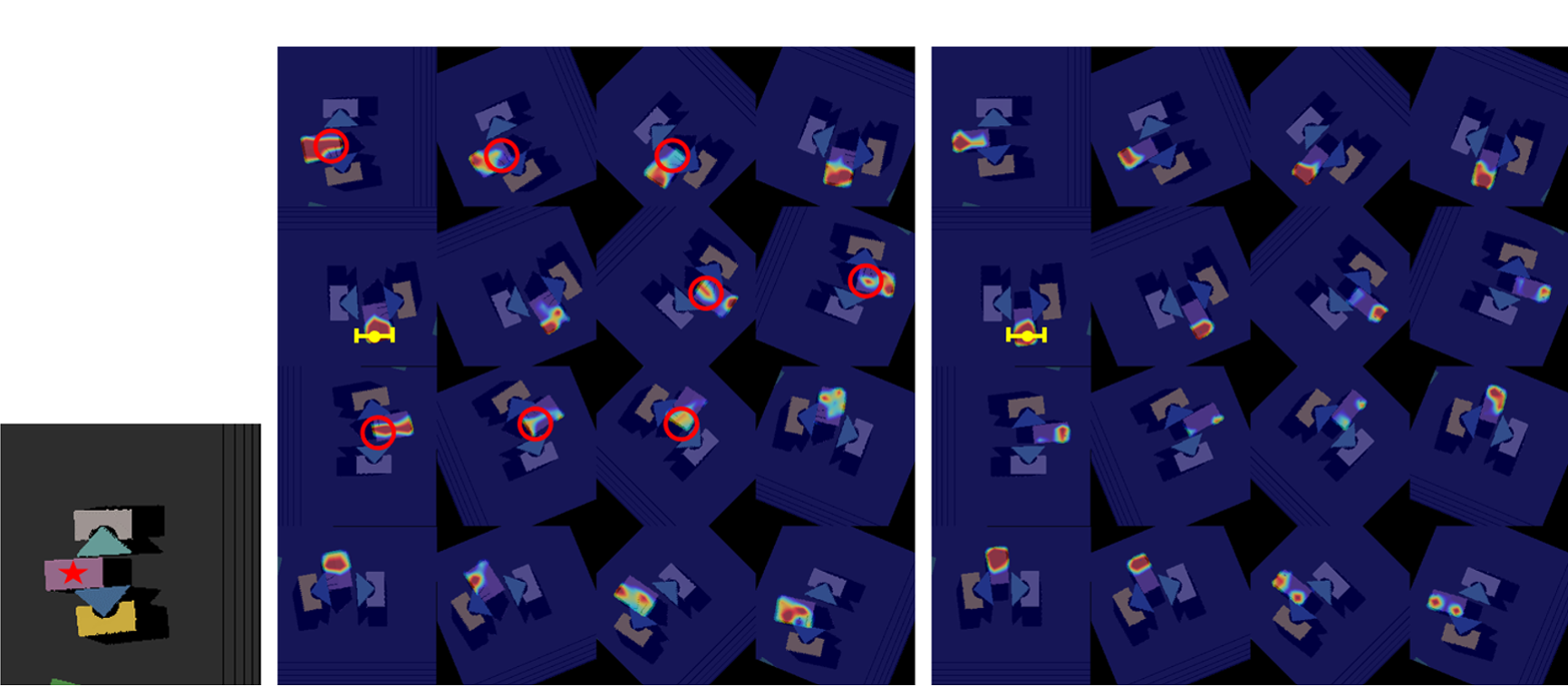}
  \vspace{-0.2cm}
  \caption{Example distributions of grasp Q values within goal object (the purple block) before (middle) and after (right) alternating training. Color tends to be red as the Q value increasing. Before alternating training, some unreasonable grasps (highlighted with red circles) show high q values. Such low quality grasps are greatly reduced after alternating training. The chosen grasp is represented by a jaw orientation and a middle point of the gripper. 
}
  \label{fig:alternating_training-1}
  \vspace{-0.6cm}
\end{figure*}

In this section we show detailed tables with metrics of each case. Real-world results of random, challenging and novel object arrangements of Table \ref{table:3} correspond to Table \ref{table:3-1}, Table \ref{table:4-1} and Table \ref{table:5-1} respectively. Simulation results of challenging arrangements of Table \ref{table:1} corresponds to Table \ref{table:2-1}. These tables show detailed comparisons and provide more comprehensive data, which shows that our method can get a better performance in the goal-oriented grasping task and is capable of generalizing to more challenging scenarios with adversarial clutter or occlusion, as well as scenarios with novel objects of more complex shape than those in training.
Also, a full version of Fig. \ref{fig:alternating_training} is shown in Fig. \ref{fig:alternating_training-1}.

% \cleardoublepage
% \newpage

\begin{table*}[t]
\caption{SIMULATION RESULTS ON CHALLENGING ARRANGEMENTS}
\setlength{\abovecaptionskip}{1cm}   
\label{table:2-1}
\vspace{-0.3cm}
\begin{center}
\scriptsize
\begin{tabular}{p{0.1\textwidth}|p{0.15\textwidth}|p{0.15\textwidth}|p{0.15\textwidth}|p{0.15\textwidth}|p{0.15\textwidth}}
\hline
 & \makecell[c]{Grasping-only} & \makecell[c]{Goal-conditioned VPG} & \makecell[c]{Grasping the Invisible} & {Our system w/o al-training} & \makecell[c]{Our system} \\
\hline
\makecell[c]{case1} & \makecell[c]{93.3 / 22.9 / 5.50} & \makecell[c]{100 / 23.8 / 6.03} & \makecell[c]{100 / 68.0 / 4.37} & \makecell[c]{100 / 95.0 / 2.07} & \makecell[c]{100 / 97.8 / 2.07} \\

\makecell[c]{case2} & \makecell[c]{63.3 / 23.8 / 5.32} & \makecell[c]{96.6 / 43.1 / 4.70} & \makecell[c]{100 / 93.3 / 2.20} & \makecell[c]{100 / 89.5 / 3.43} & \makecell[c]{100 / 83.5 / 3.40}\\

\makecell[c]{case3} & \makecell[c]{90.0 / 26.1 / 5.22} & \makecell[c]{100 / 35.5 / 4.93} & \makecell[c]{100 / 90.4 / 1.21} & \makecell[c]{100 / 92.8 / 2.27} & \makecell[c]{100 / 96.7 / 2.10} \\

\makecell[c]{case4} & \makecell[c]{23.3 / 10.8 / 18.7} & \makecell[c]{82.8 / 31.4 / 7.46} & \makecell[c]{100 / 80.0 / 5.27} & \makecell[c]{100 / 92.6 / 3.69} & \makecell[c]{100 / 98.3 / 2.03} \\

\makecell[c]{case5} & \makecell[c]{93.3 / 62.3 / 3.36} & \makecell[c]{68.4 / 84.6 / 2.79} & \makecell[c]{66.7 / 31.6 / 9.87} & \makecell[c]{76.7 / 94.4 / 3.52} & \makecell[c]{96.7 / 88.3 / 2.53} \\

\makecell[c]{case6} & \makecell[c]{100 / 32.3 / 6.33} & \makecell[c]{100 / 53.3 / 3.32} & \makecell[c]{100 / 74.2 / 3.00} & \makecell[c]{100 / 76.7 / 4.55} & \makecell[c]{100 / 83.3 / 3.40} \\

\makecell[c]{case7} & \makecell[c]{44.8 / 11.3 / 11.5} & \makecell[c]{55.2 / 32.5 / 8.46} & \makecell[c]{100 / 81.6 / 4.30} & \makecell[c]{100 / 86.3 / 3.20} & \makecell[c]{100 / 83.4 / 4.00} \\

\makecell[c]{case8} & \makecell[c]{100 / 44.8 / 2.73} & \makecell[c]{93.3 / 16.4 / 10.6} & \makecell[c]{100 / 67.1 / 3.02} & \makecell[c]{100 / 84.2 / 3.13} & \makecell[c]{100 / 95.0 / 2.17} \\

\makecell[c]{case9} & \makecell[c]{86.7 / 21.7 / 11.5} & \makecell[c]{100 / 74.7 / 2.90} & \makecell[c]{83.3 / 63.3 / 6.90} & \makecell[c]{100 / 81.0 / 3.87} & \makecell[c]{93.3 / 78.5 / 3.92} \\

\makecell[c]{case10} & \makecell[c]{93.3 / 34.1 / 4.61} & \makecell[c]{96.7 / 21.5 / 6.58} & \makecell[c]{100 / 54.5 / 3.15} & \makecell[c]{100 / 83.3 / 2.30} & \makecell[c]{100 / 96.7 / 2.07} \\

\makecell[c]{average} & \makecell[c]{78.8 / 29.0 / 7.48}  & \makecell[c]{89.3 / 41.7 / 5.78} & \makecell[c]{95.0 / 70.4 / 4.33} & \makecell[c]{{97.7} / {87.6} / {3.20}} & \makecell[c]{{\bf 99.0} / {\bf 90.2} / {\bf 2.77}}\\
% \hline
% \hline
% \makecell[c]{random} & \makecell[c]{90.0 / 60.3 / 4.20}  & \makecell[c]{93.8 / 62.3 / 3.93} & \makecell[c]{96.7 / 54.6 / 5.37} & \makecell[c]{97.8 / 83.7 / 4.50} & \makecell[c]{{\bf 97.8} / {\bf 90.0} / 4.82}\\
\hline
\end{tabular}
\end{center}
\begin{tablenotes}
\scriptsize
     \item * Number of each cell is presented as “Completion / Grasp Success / Motion Number” to display the detailed testing performance of all the cases in Figure \ref{fig:case}.
\end{tablenotes}
\vspace{-0.5cm}
\end{table*}

\end{document}